%% file: main_arxiv.tex
\documentclass[10pt,twocolumn,letterpaper]{article}

\usepackage[pagenumbers]{cvpr}
\usepackage{graphicx}
\usepackage{amsmath}
\usepackage{amssymb}
\usepackage{enumitem}
\usepackage{multirow}
\usepackage{booktabs}
\usepackage{mathtools, nccmath}
\usepackage{siunitx}
\usepackage{color, colortbl}
\usepackage{etoolbox}
\AtBeginEnvironment{quote}{\par\singlespacing\small}
\usepackage[dvipsnames]{xcolor}
\usepackage[nohyperlinks]{acronym}
\usepackage{hyperref}
\hypersetup{pagebackref,breaklinks,colorlinks}
\usepackage[capitalize]{cleveref}
\crefname{section}{Sec.}{Secs.}
\Crefname{section}{Section}{Sections}
\Crefname{table}{Table}{Tables}
\crefname{table}{Tab.}{Tabs.}
\newcommand\Mark[1]{\textsuperscript#1} 

\acrodef{dnn}[DNN]{Deep Neural Network}
\acrodef{se}[SE]{Squeeze and Excitation}
\acrodef{bn}[BN]{Batch Normalization}
\acrodef{in}[IN]{Instance Normalization}
\acrodef{relu}[ReLU]{Rectified Linear Unit}
\acrodef{prelu}[PReLU]{Parametric \acs{relu}}
\acrodef{gelu}[GELU]{Gaussian Error Linear Unit}
\acrodef{silu}[SiLU]{Sigmoid Linear Unit}
\acrodef{psilu}[PSiLU]{Parametric \acs{silu}}
\acrodef{pssilu}[PSSiLU]{Parametric Shifted \acs{silu}}

\acrodef{gmac}[GMAC]{Giga Multiply–Accumulate}
\acrodef{pp}[pp]{percentage points}

\acrodef{xcit}[XCiT]{Cross-Covariance Image Transformers}
\acrodef{vit}[ViT]{Vision Transformer}
\acrodef{deit}[DeiT]{Data-efficient image Transformers}

\acrodef{preact}[PreAct]{Pre-Activation}

\acrodef{at}[AT]{Adversarial Training}
\acrodef{pgd}[PGD]{Projected Gradient Descent}
\acrodef{aa}[AA]{AutoAttack}
\acrodef{fgsm}[FGSM]{Fast Gradient Sign Method}

\acrodef{ra}[RobArch]{Robust Architecture}

\acrodef{sota}[SOTA]{state-of-the-art}

\DeclareMathOperator*{\argmin}{argmin}

\mathchardef\mh="2D
\DeclarePairedDelimiter{\nint}\lfloor\rceil

\definecolor{mygreen}{RGB}{93, 150, 122}
\definecolor{mypurple}{RGB}{121, 120, 173}
\definecolor{mygray}{RGB}{50, 50, 50}
\definecolor{myorange}{HTML}{fee6ce}
\definecolor{mylightgray}{HTML}{f0f0f0}

\newcommand{\mcolor}[2]{\textcolor{#1}{#2}}

\begin{document}

\title{RobArch: Designing Robust Architectures against Adversarial Attacks}

\author{
ShengYun Peng\Mark{1}, Weilin Xu\Mark{2}, Cory Cornelius\Mark{2}, Kevin Li\Mark{1}, Rahul Duggal\Mark{1}, \\
Duen Horng Chau\Mark{1}, and Jason Martin\Mark{2} \\
\Mark{1}Georgia Institute of Technology, Atlanta, GA, USA\\
{\tt\small \{speng65,kevin.li,rahulduggal,polo\}@gatech.edu} \\
\Mark{2} Intel Corporation, Hillsboro, OR, USA\\
{\tt\small \{weilin.xu,cory.cornelius,jason.martin\}@intel.com} \\
}
\maketitle

\begin{abstract}
\acl{at} is the most effective approach for improving the robustness of \acp{dnn}.
However, compared to the large body of research in optimizing the adversarial training process, there are few investigations into how architecture components affect robustness, and they rarely constrain model capacity.
Thus, it is unclear where robustness precisely comes from. 
In this work, we present the first large-scale systematic study on the robustness of \ac{dnn} architecture components under fixed parameter budgets.
Through our investigation, we distill 18 actionable robust network design guidelines
that empower model developers to gain deep insights.
We demonstrate these guidelines' effectiveness by introducing the novel \ac{ra} model that instantiates the guidelines to build a family of top-performing models across parameter capacities against strong adversarial attacks.
\ac{ra} achieves the new \acl{sota} \acl{aa} accuracy on the RobustBench ImageNet leaderboard.
The code is available at \href{https://github.com/ShengYun-Peng/RobArch}{https://github.com/ShengYun-Peng/RobArch}.

\end{abstract}

\section{Introduction}
\label{sec:intro}

\acfp{dnn} are vulnerable to adversarial attacks~\cite{goodfellow2014explaining, szegedy2013intriguing, kurakin2016adversarial, liu2018dpatch, brown2017adversarial}.
Many defense methods have been proposed to mitigate this pitfall~\cite{andriushchenko2020understanding, zhang2019theoretically, xie2017mitigating, song2017pixeldefend, xie2019feature, tu2020physically, das2022skelevision},
and among them, \acf{at}~\cite{madry2018towards} is the most effective way to defend against adversarial attacks. 
Compared to the large body of research devoted to improving the loss function~\cite{hosseini2021dsrna, liu2020loss} and optimizing the \ac{at} procedure~\cite{zhang2019theoretically, wong2020fast, ding2018mma}, few studies investigate how architectural components affect robustness despite its importance. 

Yet \ac{dnn} architectures have been dominating generalization improvements~\cite{he2016deep, dosovitskiy2020image, liu2022convnet}.
Recent research has started to highlight the potential significant impact architecture choices could have on robustness \cite{su2018robustness,devaguptapu2021adversarial}, and showed that adjusting widths~\cite{wu2021wider} or depths~\cite{huang2021exploring} could robustify a network.

\begin{figure}[t]
  \centering
   \includegraphics[width=0.9\linewidth]{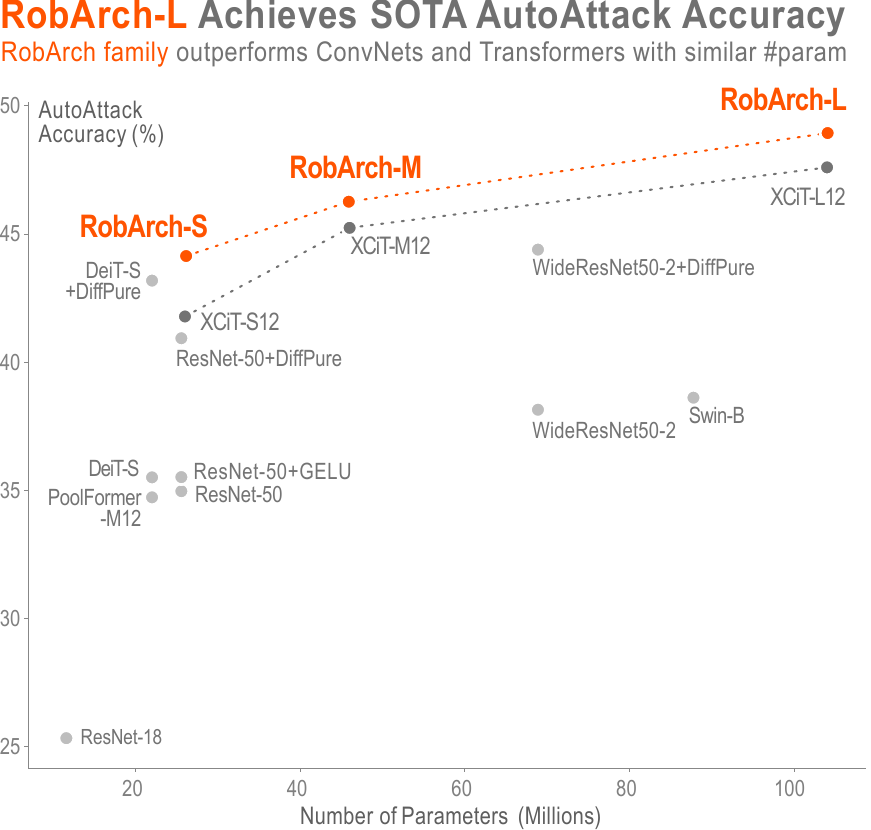}
   \caption{
   Our \acs{ra} model family outperforms the \acs{sota} \acs{xcit} family on RobustBench ImageNet leaderboard~\cite{croce2020robustbench}.
   Every \acs{ra} model outperforms its \acs{xcit} counterparts at a similar capacity.
   \acs{ra}-S outperforms ResNet-50 by 9.18 percentage points, and is even more robust than WideResNet50-2 despite having 2.6$\times$ fewer parameters. The robustness continues to increase as capacity increases. \ac{ra}-L achieves the new \acs{sota} \acs{aa} accuracy on RobustBench. Table \ref{tab:sota_paper} presents accuracy details.}
   \label{fig:aa-sota}
\end{figure}

However, those studies did not constrain the model capacity, making it hard to attribute the robustness gains to those adjustments, because increasing model capacity alone could already improve robustness~\cite{madry2018towards,huang2021exploring}. 
Thus, controlling for model capacity while assessing robustness is important, and recent research has provided supporting evidence.
For example, despite the popular belief that transformer models might be more robust than CNNs~\cite{bhojanapalli2021understanding, shao2021adversarial}, Bai \etal~\cite{bai2021transformers} demonstrated that \ac{deit}~\cite{touvron2021training} and ResNet~\cite{he2016deep} with \ac{gelu} activations~\cite{hendrycks2016gaussian} attained comparable robustness if the model scales were balanced. 
Therefore, it remains unclear how these previously studied architectural components precisely affect robustness.
Our research filled this critical research gap by making three key contributions:

\begin{itemize}[leftmargin=*,topsep=0pt]
\itemsep-0.1em 
\item \textbf{The first large-scale systematic study on the robustness of \ac{dnn} architecture components.} 
To the best of our knowledge, our work is the first to comprehensively investigate and compare the robustness impacts of a wide range of architecture components on a large dataset such as ImageNet.
Advancing over prior work, we carefully constrain the parameter budget to isolate and hone in on the benefit of each component.
Such a systematic study enables us to discover a family of new architectures that outperform \ac{sota} networks. (Figure \ref{fig:aa-sota}). 

\item \textbf{18 actionable robust network design guidelines.}
Our systematic investigation for component robustness, 
through training over 150 models on ImageNet~\cite{deng2009imagenet},
enables us to distill 18 generalizable, actionable guidelines that empower model developers to gain deep insights and design networks with higher robustness.
The guidelines present significant new knowledge and discoveries for our computer vision community.
For example, we have discovered  
(1) deepening a network is more effective than widening it, and there is a sweet spot; 
(2) specific modifications such as adding \ac{se} block, removing the first normalization layer in a block, and reducing the downsampling factor in the stem stage effectively boosts robustness; and
(3) architecture designs that harm robustness include inverted bottleneck, large dilation factor, \ac{in}, parametric activation functions~\cite{dai2022parameterizing}, and reducing activation layers. 

\item \textbf{Top performance against strong adversarial attacks.} 
We demonstrate our guidelines' effectiveness by introducing the novel \acf{ra} model that instantiates the guidelines to build a family of top-performing models across parameter capacities against strong adversarial attacks.
In particular, we compare our \ac{ra} family with the \acf{xcit} family~\cite{ali2021xcit} that is the \ac{sota} on RobustBench~\cite{croce2020robustbench}. 
Every \ac{ra} model
outperform its \ac{xcit} counterpart with a similar model capacity (Figure \ref{fig:aa-sota}).
\ac{ra}-S surpasses ResNet-50's \ac{aa} accuracy by 9.18 percentage points, and is even more robust than WideResNet50-2 despite having 2.6$\times$ fewer parameters. 
The robustness continues to increase as capacity increases.
\ac{ra}-L achieves the new \ac{sota} \acf{aa}~\cite{croce2020reliable} accuracy 
on the RobustBench ImageNet leaderboard.
\ac{ra}'s performance advantage extrapolates to the \ac{pgd} attack. 
Overall, the proposed \acp{ra} outperform both ConvNets and Transformers with similar total parameters. 

\end{itemize}

\section{Robust Architecture Design}
\label{sec:robust_arch_design}

We carefully select architectural components from off-the-shelf \acp{dnn} (ResNet~\cite{he2016deep}, RegNet~\cite{radosavovic2020designing}, DenseNet~\cite{huang2017densely}, and ConvNeXt~\cite{liu2022convnet}) that improve generalization accuracy.
Based on the commonalities in these network designs, we group the components into three modification categories:
\begin{itemize}[label={\tiny$\bullet$}, leftmargin=12pt,topsep=0pt]
\itemsep-0.2em 
    \item \textbf{Network-level}: \textit{depth}, \textit{width}
    \item \textbf{Stage-level}: \textit{stem stage}, \textit{dense connection}
    \item \textbf{Block-level}: \textit{kernel size}, \textit{dilation}, \textit{activation}, \textit{\ac{se}}, \textit{normalization}
\end{itemize}
Since ResNet \cite{he2016deep} is a milestone in the history of \ac{dnn} architecture, we choose its most popular instantiation, ResNet-50 (${\sim}26$M parameters) as the base architecture, which consists of a stem stage, $n=4$ body stages, and a classifier head, as our starting point. 
Each body stage contains multiple residual blocks with various depth and width configurations. 
Appendix \ref{sec:appendix-network} provides details of ResNet-50 configurations. 

\smallskip 
\noindent \textbf{Notation and symbols used throughout this paper.} 

\begin{itemize}[topsep=0pt, itemsep=0mm, parsep=1mm, leftmargin=9pt]

\item We denote $D\mh d_1 \mh ... \mh d_n$ as the depth of each stage in an $n$-stage network $\left( n \in \{ 3, 4, 5, 6 \} \right) $. 

\item For stage $i$, $w_i$ and $w_{b_i}$ are the numbers of channels in the pointwise and non-pointwise convolutions, respectively. 

\item Bottleneck multiplier $b_i$ is the ratio of channels in pointwise to non-pointwise convolution, $b_i = w_i / w_{b_i}$.

\item Assuming $w_{g_i}$ is the group convolution width, $g_i$ is the total number of groups in the non-pointwise convolution layer:  
$g_i = \nint{w_{b_i} / w_{g_i}} = \nint{w_i / \left( b_i \times w_{g_i} \right) }$. 

\item Width expansion ratio is $e = w_{i + 1} / w_i, \ i \leq n - 1$. 

\item We use $W \mh w_1 \mh ... \mh w_n$, $G \mh g_1 \mh ... \mh g_n$, $BM \mh b_1 \mh ... \mh b_n$ to represent the number of channels, group convolution groups, and bottleneck multiplier in an $n$-stage network.

\end{itemize}

\smallskip
\noindent \textbf{Experimental settings.} We train all models on ImageNet~\cite{deng2009imagenet} with the recipes specified in Sec. \ref{sec:training}. 
When studying a single architecture component (Sec. \ref{sec:network} - \ref{sec:block}) and building cumulative networks (Sec. \ref{sec:roadmap} \& \ref{sec:scale_up}), we use 10-step \ac{pgd} (\ac{pgd}10) with different attack budgets $\epsilon$ $\left( \epsilon \in \{2, 4, 8 \} \right)$ for fast evaluations. 
After finalizing the model structures of the \ac{ra}, we test all \acp{ra} against \ac{pgd}100 and \ac{aa}. 
All attacks are $\ell_\infty$ bounded. 
To control for the effect of model capacity,
we constrain the networks' total parameters, \ie, similar to ResNet-50 (${\sim} 26$M), throughout the exploration. 

\begin{figure*}
\centering
    \begin{subfigure}{0.34\linewidth}
        \includegraphics[height=1.9in]{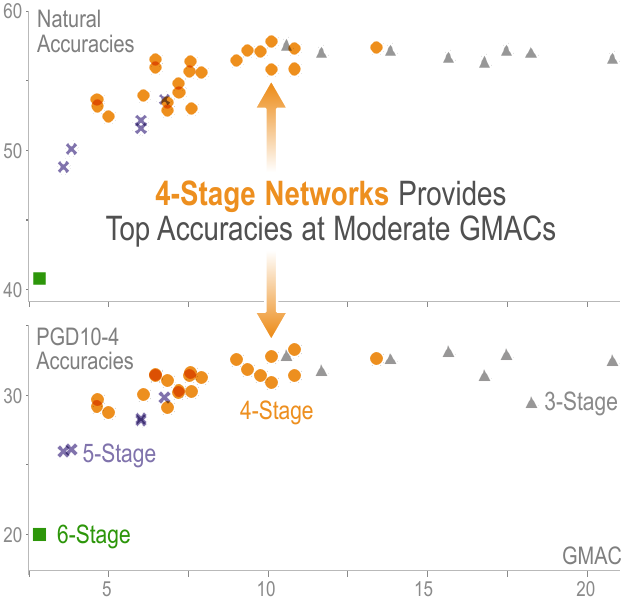}
        \caption{
        4-stage networks attain top accuracies at much lower \acp{gmac} than 3-stage networks.
        5-stage and 6-stage networks are significantly less robust. 
        }
        \label{fig:depth-stages}
    \end{subfigure}
    \hfill
    \begin{subfigure}{0.29\linewidth}
        \includegraphics[height=1.9in]{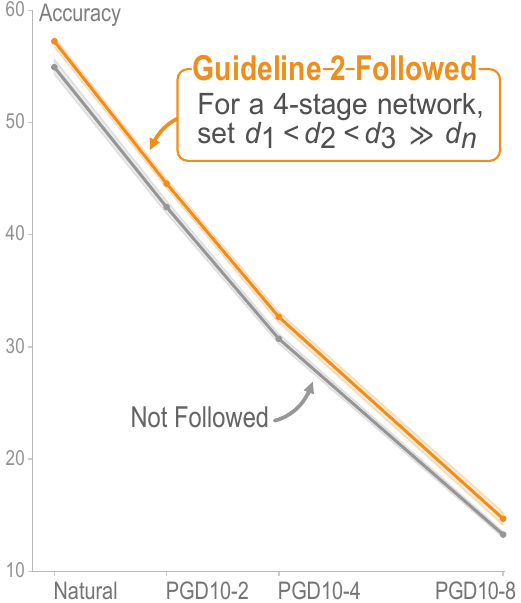}
        \caption{Higher accuracy when guideline 2 is followed: depth rule $d_1 < d_2 < d_3 > c \times d_4$. We plot the mean accuracy (solid line) and $95\%$ confidence interval (the shading). 
        }
        \label{fig:depth-rule}
    \end{subfigure}
    \hfill
    \begin{subfigure}{0.32\linewidth}
        \includegraphics[height=1.9in]{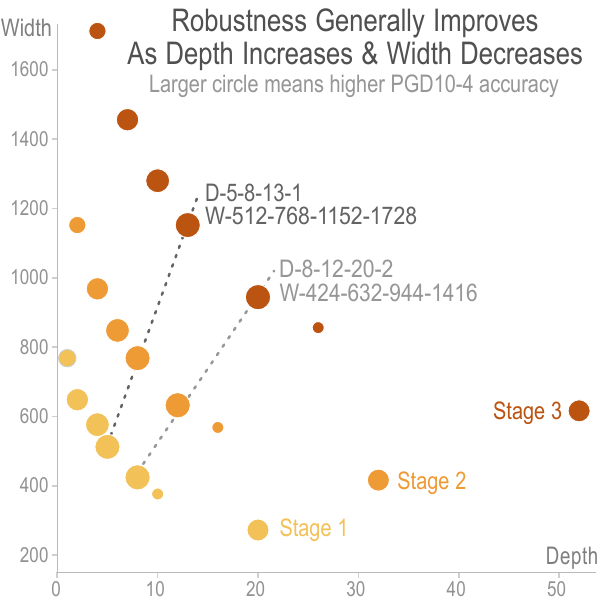}
        \caption{
        Robustness generally improves in all stages as depth increases and width decreases,
        until catastrophic overfitting happens with significantly reduced  robustness.}
        \label{fig:depth-width}
    \end{subfigure}
\caption{For network-level design, following guideline 2 to increase depth and decrease width in a 4-stage network provides optimal robustness. 
We study (a) how the number of stages affects accuracies, (b) stage depth settings, and (c) depth-width trade-off.  
We only plot the first three stages of a 4-stage network in (c) for better visualization since the last stage is much shallower as per the optimal depth configurations in guideline 2. 
These observations also apply to other \ac{pgd} attack budgets, as shown in Appendix \ref{sec:appendix-arch-components}.}
\label{fig:depth}
\end{figure*}

\subsection{Training Techniques}
\label{sec:training}

\noindent \textbf{Standard-\ac{at}.} 
\acf{at} is the most reliable defense to obtain robust \acp{dnn}~\cite{madry2018towards, goodfellow2014explaining}. 
Standard-\ac{at} is formulated as a min-max optimization framework~\cite{madry2018towards}.
Given a network $f_\theta$ parameterized by $\theta$, a dataset with samples $(x_i, y_i)$, and a loss function $\mathcal{L}$, the robust optimization problem is formulated as:
\begin{equation}
    \argmin_\theta \mathbb{E}_{(x_i, y_i) \sim \mathbb{D}} \left[ \max_{x'} \mathcal{L} \left( f_\theta, x', y \right) \right],  
\end{equation}
The inner adversarial example $x'$ is generated on the fly during the training process, which aims to find an adversarial perturbation of a given data point $x$ that achieves a high loss, 
\begin{equation}
    x'_{k + 1} = \prod_{x+\Delta} \left( x'_k + \alpha \text{sgn} \left( \nabla_x \mathcal{L}(\theta, x'_k, y) \right) \right).
\end{equation}
$\text{sgn}(\cdot)$ is the sign function, $\alpha$ is the step size, $x'_k$ is the adversarial example generated after $k$ steps ($1 \leq k \leq K$), $\Delta = \{ \delta: \| \delta \|_\infty \leq \epsilon \}$ is the threat mode, and $\prod_{x+\Delta}$ is a projection operation that clips the perturbation back to the $\epsilon$-ball centered on $x$  if it goes beyond the attack budget.

\noindent \textbf{Fast-\ac{at}.}
Fast-\ac{at} speeds up the Standard-\ac{at} and can robustify a ResNet-50 in under 13 hours~\cite{wong2020fast}. 
It not only adopts \ac{fgsm}~\cite{goodfellow2014explaining} to generate adversarial samples during the training but also incorporates a cyclic learning rate~\cite{smith2017cyclical} and mixed-precision arithmetic~\cite{micikevicius2017mixed} to fully accelerate the \ac{at} with just 15 epochs. 
A line of research improves the performance and mitigates the catastrophic overfitting problem discovered in the Fast-\ac{at}, \eg, YOPO~\cite{zhang2019you}, GradAlign~\cite{andriushchenko2020understanding}, GAT~\cite{sriramanan2020guided}, Sub-\ac{at}~\cite{li2022subspace}, \etc, but there are limited explorations on whether these recipes are compatible with the full ImageNet~\cite{deng2009imagenet}.

Although Fast-\ac{at} provides competitive \ac{pgd} results, its resulting robustness on ResNet-50 is inferior to that of Standard-\ac{at}'s as per the \ac{aa} accuracy on the RobustBench leaderboard~\cite{croce2020robustbench}. 
Therefore, we use Fast-\ac{at} as a rapid indicator while exploring different architecture components and building the \ac{ra} family, and use Standard-\ac{at} to robustify all members in the \ac{ra} family.

\subsection{Network-level Design}
\label{sec:network}

\noindent \textbf{Depth.}
In the standard ResNet-50 ($D\mh 3 \mh 4 \mh 6 \mh 3$), each stage downsamples the input features by 2. 
The downsampling in the first stage is replaced by a max-pooling layer in the stem stage. 
We sample 36 architectures based on the depth relationship between each pair of stages, \ie, $d_i \leq d_{i + 1}$ and $d_i > d_{i + 1}$. 
The widths in all stages are the same as ResNet-50, and when $n > 4$, we reuse the width in stage 4. 
For $n = 6$, even setting $d_i = 1, i \leq n$ leads to $1.83$M more parameters than ResNet-50. 
Hence, there is only 1 data point for the 6-stage network, and we do not continue increasing the total stages.
Fig. \ref{fig:depth-stages} shows the results after \ac{at}. 
4-stage networks attain top natural and adversarial accuracies at much lower \acp{gmac} than 3-stage networks.
5-stage and 6-stage networks are significantly less robust. 
These results are expected since shallow stages, in general, compute on higher resolutions, and the depth of a 3-stage network in shallow stages is deeper than a 4-stage network by a large margin for similar total parameters. 
Hence, we select 4-stage networks and further explore the depth relationship between stages. 

Huang \etal~\cite{huang2021exploring} found that reducing depth in the last stage of a 3-stage WideResNet34-10 improves robustness. 
Upon further inspection of our 4-stage models, we observe that increasing the stage depths $d_i$ along with $i$, then significantly decreasing the depth in the last stage, leads to higher robustness. 
Fig. \ref{fig:depth-rule} shows
that following such a rule ($d_1 < d_2 < d_3 > c \times d_4$) leads to a higher accuracy than not following it.
We set $c = 3$ and leave the finetuning of a larger $c$ to further research. 
RegNet~\cite{radosavovic2020designing} first discovered the depth pattern and applied it to improve benign accuracy.
Our results extend this discovery to adversarial settings and show that it helps robustify architectures without incurring extra parameters.
Overall, we found the optimal stage depth ratio is $D \mh 5 \mh 8 \mh 13 \mh 1 $ and listed its performance in Table \ref{tab:components} row 2.

\textbf{Guideline 1:} \textit{3-stage $\approx$ 4-stage $>$ 5-stage $\gg$ 6-stage network in terms of robustness.}

\textbf{Guideline 2:} \textit{For a 4-stage network, set $d_1 < d_2 < d_3 \gg d_n$, and $D \mh 5 \mh 8 \mh 13 \mh 1 $ provides the optimal robustness. }

\smallskip
\noindent \textbf{Width.} 
Factors that affect the stage width are pointwise convolution channels $w_i$, group convolution groups $g_i$, and bottleneck multiplier $b_i$.
The width configurations of the standard ResNet-50 are $W \mh 256 \mh 512 \mh 1024 \mh 2048$, $G \mh 1 \mh 1 \mh 1 \mh 1 $, $BM \mh 0.25 \mh 0.25 \mh 0.25 \mh 0.25$.
Unless otherwise specified, all configurations are kept consistent with ResNet-50 when studying one of the factors. 

For $b_i \in \{0.125, 0.25, 0.5, 1, 2, 4\}$, we first test a constant $b_i = b$ in all stages.
The accuracy reaches the peak when $b = 0.25$ or $0.5$ and significantly decreases when increasing $b$ from $0.5$ to $4$, which shows the inverted bottleneck is harmful to robustness. 
$b = 0.25$ (ResNet-50) has higher natural and \ac{pgd}10-2 accuracy, while $b = 0.5$ has higher \ac{pgd}10-4 and \ac{pgd}10-8 accuracy. 
Both results are shown in Table \ref{tab:components} (rows 1 and 3). 
Then, we vary $b_i$ for different stages, $b_{1,2} < b_{3,4}$ and $b_{1,2} > b_{3,4}$.
The robustness of $BM \mh 0.25 \mh 0.25 \mh 2 \mh 2$ is better than $b_i = 2$ but worse than $b_i = 0.25$. 
Surprisingly, $BM \mh 4 \mh 4 \mh 0.25 \mh 0.25$ outperforms both $b_i = 0.25$ and $b_i = 4$. 
We further combine the two optimal bottleneck multipliers and set $b_{1,2} = 0.5, b_{3,4} = 0.25$.
As shown in Table \ref{tab:components} row 4, this setting attains higher accuracy than both $b_i = 0.5$ and $0.25$. 

Next, we study the group convolution groups $g_i \in \{ 1, 2, 4, 8, 16, w_{b_i} \}$. 
$g_i = w_{b_i}$ is equivalent to the depth convolution. 
The pointwise convolution width $w_i$ is adjusted to reach the controlled parameter budget, but $b_i$ is always $0.25$. 
For a constant $g_i = g$, we observe a significant increase from $g = 1$ (ResNet-50) to $g = 2$, but then the accuracy gradually decreases if we continue to increase $g$. 
Similar to the bottleneck multiplier study, we vary $g_i$ for different stages. 
However, there is no further robustness gain.
We list the results of $g = 2$ in Table \ref{tab:components} row 5.

For the width expansion ratio, we evaluate $e \in \{1, 1.5, 2, 2.5, 3 \}$.
The robustness rises and saturates at $e = 1.5$ and falls for a larger $e$.
We show $e = 1.5$ in Table \ref{tab:components} row 6. 
Finally, we combine the optimal configurations for all three factors, \ie, $b_{1,2} = 0.5, b_{3,4} = 0.25, g_i = g = 2, e = 1.5$.
However, the robustness is inferior to that of just using the individual optimal settings. 
After a close look at all the results, we find setting a constant $b_i = b = 0.25$ works favorably with $g$ and $e$. 
In addition, we observe $g = 2, e = 2$ and $g = 1, e = 1.5$ achieve the best two accuracies. 
The phenomenon also demonstrates that directly combining multiple individual optimal architectural settings does not transfer to a better model.

\textbf{Guideline 3:} \textit{Inverted bottleneck harms robustness, especially when added to deeper stages.}

\textbf{Guideline 4:} \textit{For a single modification, $b_{1,2} = 0.5, b_{3,4} = 0.25$, $g_i = 2$, and $e = 1.5$ all show promising improvements. However, merging all three configurations makes the model less robust, and the optimal width configurations are $e = 2, g = 2$ or $e = 1.5, g = 1$ with $b = 0.25$.}

\smallskip
\noindent \textbf{Combining Depth and Width.}
In this part, we answer the following question: \textit{Under a fixed model capacity, does increasing widths while decreasing depths, or vice versa, improve robustness?}

We use the optimal depth ratio, $D \mh 5 \mh 8 \mh 13 \mh 1$. 
To provide a more general understanding and avoid overfitting to specific optimal settings, we cross-select $e = 1.5, g = 2, b = 0.25$ from the two optimal width configurations from guideline 4. 
We proportionally adjust depths and widths to accommodate the fixed budget. 
Fig. \ref{fig:depth-width} displays the relationship between depths and widths using \ac{pgd}10 accuracy. 
A larger bubble size means higher accuracy. 
The results show that increasing depth while decreasing width improves robustness in all stages.
It is important to note that if we continue the trend, catastrophic overfitting~\cite{andriushchenko2020understanding} occurs during training.
Since catastrophic overfitting drastically decreases the robustness, we should deepen the network but balance the depth and the width to stabilize the \ac{at} process.  
Comparing the top 2 models (dotted lines), both \ac{pgd}10-2 and \ac{pgd}10-4 accuracies of the deeper model are $0.10pp$ (percentage points) higher, but the \ac{pgd}10-8 accuracy is $0.49pp$ lower, which is a sign of unstable training. 
Overall, $D \mh 5 \mh 8 \mh 13 \mh 1$ is selected as the starting point of our cumulative model in Sec. \ref{sec:roadmap}. 
Compared to ResNet-50 ($D \mh 3 \mh 4 \mh 6 \mh 3$), $D \mh 5 \mh 8 \mh 13 \mh 1$ is much deeper and slimmer with significantly higher robustness: 
$\uparrow1.15pp$  for natural accuracy, $\uparrow2.03pp$ for \ac{pgd}10-2, $\uparrow2.62pp$ for \ac{pgd}10-4, and $\uparrow2.75pp$ for \ac{pgd}10-8.
We observe a similar depth-width relationship when scaling up the model in Sec. \ref{sec:scale_up}. 

\textbf{Guideline 5:} \textit{Under a fixed model capacity, first increase the network depth proportionally to the optimal depth until catastrophic overfitting happens, \ie, a sudden drop in loss and increase in training accuracy. The width is adjusted to fill the total parameter budget. } \par

\begin{table}
\small
\centering
\caption{\ac{pgd}10 robustness of architecture components. All configurations trained with Fast-\ac{at} and evaluated on full ImageNet validation set. We provide ResNet-50 as baseline.  
Appendix \ref{sec:appendix-arch-components} shows detailed results, including \ac{pgd}10-2 and \ac{pgd}10-8.}
\begin{tabular}{rlrr}
\toprule
    Idx. & Configurations & Natural &  PGD10$\mh$4 \\
\midrule
    1 & ResNet-50 & 56.09\% & 30.43\% \\
\midrule
  \multicolumn{4}{c}{\textit{Network-level Design}}\\
\midrule
    2 & $D\mh 5 \mh 8 \mh 13 \mh 1$ & 57.35\% & 33.33\% \\
    3 & $BM\mh 0.5 \mh 0.5 \mh 0.5 \mh 0.5 $ & 55.31\% & 30.52\% \\
    4 & $BM \mh 0.5 \mh 0.5 \mh 0.25 \mh 0.25$ & 56.11\% & 31.26\% \\
    5 & $G \mh 2\mh 2\mh 2\mh 2$ & 57.31\% & 32.09\% \\
    6 & $W \mh 512 \mh 768 \mh 1152 \mh 1728$ & 57.17\% & 32.04\% \\
    \multirow{3}{*}{7} & $G\mh 2 \mh 2\mh 2\mh 2$ & \multirow{3}{*}{56.64\%} & \multirow{3}{*}{31.04\%} \\
      & $BM\mh 05\mh 05\mh 025\mh 025$ & & \\
      & $W\mh 512\mh 768\mh 1152\mh 1728$ & & \\
\midrule
  \multicolumn{4}{c}{\textit{Stage-level Design}}\\
\midrule
    8 & Stem width 96 & 57.29\% & 32.06\% \\
    9 & Move down ($\downarrow$) downsampling & 57.08\% & 33.08\% \\
    10 & Dense ratio 2 & 55.93\% & 30.73\% \\
\midrule
  \multicolumn{4}{c}{\textit{Block-level Design}}\\
\midrule
    11 & Kernel size 5 & 56.73\% & 32.77\% \\
    12 & Kernel size 7 & 59.70\% & 34.67\% \\
    13 & Dilation 2 & 52.98\% & 28.38\% \\
    14 & Dilation 3	& 52.10\% & 27.97\% \\
    15 & Act. \acs{gelu} & 57.48\% & 33.12\% \\
    16 & Act. \acs{silu} & 58.19\% & 34.07\% \\
    17 & Act. \acs{psilu} & 56.38\% & 33.76\% \\
    18 & \ac{se} (\acs{relu}) & 57.83\% & 32.64\% \\
    19 & Norm$\mh$BN$\mh$BN$\mh$0 & 54.15\% & 29.59\% \\
    20 & Norm$\mh$BN$\mh$0$\mh$BN & 56.04\% & 31.34\% \\
    21 & Norm$\mh$0$\mh$BN$\mh$BN & 56.18\% & 31.61\% \\
\bottomrule
\end{tabular}
\label{tab:components}
\end{table}

\subsection{Stage-level Design}
\label{sec:stage}

\noindent \textbf{Stem Stage.} 
The stem stage in a standard ResNet-50 consists of a convolution layer and a max-pooling layer, each of which has a downsampling factor of 2. 
All 4 tandemly-connected body stages downsample the input resolution by 2 except the first stage.
The convolution layer uses a $7 \times 7$ kernel and outputs 64-layer features. 

In the stem stage, we modify the following architectural components: channel width, kernel size, ``patchify'' stem, and downsampling factor. 
First, we test channel width $\in \{ 32, 64, 96\}$ and kernel size $\in \{ 3, 5, 7 \}$.
With less than $0.01$M increase in total parameters, switching convolution layer width from $32$ to $64$ and $64$ to $96$ improve the \ac{pgd}10-4 accuracy by $0.7$ and $1.65$ percentage points, respectively.
The ``stem width 96'' is located in Table \ref{tab:components} row 8. 
For kernel size $= 3$ or $=5$, the training overfits to \ac{fgsm} and leads to a completely non-robust model.
The original kernel size is $7$ in ResNet-50, and increasing it to $9$ improves the \ac{pgd} accuracy but leads to a drop in the natural accuracy.

We study the downsampling factor next. 
RegNet~\cite{radosavovic2020designing} is built based on ResNet, but the max-pooling layer in the stem stage is replaced by a stride 2 convolution shortcut connection in the first stage. 
We denote this operation as ``move down ($\downarrow$) downsampling.''
The evaluation result (Table \ref{tab:components} row 9) manifests $0.99$ and $2.65$ percentage points increments in natural and \ac{pgd}10-4 accuracy. 
We further disassemble the operation by only discarding the max-pooling layer without adding the stride 2 convolution shortcut.
Although the robustness is slightly lower than ``move $\downarrow$ downsampling,'' it still outperforms ResNet-50 by a large margin. 

\ac{vit}~\cite{dosovitskiy2020image} first introduced the ``patchify stem,'' and ConvNeXt~\cite{liu2022convnet} also incorporated the design to improve generalization.
Motivated by those works, we replace the original stem with a $4 \times 4$ patch, \ie, kernel size $=$ stride $= 4$, and observe a slight increment in robustness. 
Since moving down the downsampling layer boosts robustness, we continue to test a smaller $2 \times 2$ patch. 
The accuracy increases as expected, but the gain is slightly lower than directly moving down the downsampling layer in a ResNet-style stem. 
Since a small kernel size in the early convolution layer leads to a smaller receptive field, a moderate kernel size of $7 \times 7$ is preferred. 
Overall, we select ``stem width 96'' and ``move $\downarrow$ downsampling'' as potential candidates while building the cumulative model in Sec. \ref{sec:roadmap}.

\textbf{Guideline 6:} \textit{Replacing the max-pooling in the stem stage with a downsampling shortcut in the first stage significantly improves robustness.}

\textbf{Guideline 7:} \textit{For the convolution layer in the stem stage, directly replacing it with a ``patchify'' stem design contributes to the robustness. However, the optimal configurations are increasing the channel width and setting kernel size $=7$.} \\

\smallskip
\noindent \textbf{Dense Connection.}
Huang \etal~\cite{huang2017densely} introduced the dense connection in DenseNet that concatenates the feature maps of all preceding blocks within the stage as the input to the current block.
We extend the definition and experiment with different dense ratios $i$ $\left( i \in \{1, 2, 3, 4, 5\} \right)$, \ie, $i$ preceding feature maps are used to construct the input. 
Only $i = 2$ shows minor improvements in \ac{pgd} accuracy, and no strong benefits are observed (Table \ref{tab:components} row 10). 
We further remove the last \ac{relu} since the original DenseNet uses the \ac{preact} operation~\cite{he2016identity}.
However, the robustness is further degraded, and we assume the poor performance of reducing the last activation itself (discussed in \ref{sec:block}) is a potential reason. \par

\textbf{Guideline 8:} \textit{Dense connection is not beneficial to robustness.}

\subsection{Block-level Design}
\label{sec:block}

\noindent \textbf{Kernel Size.}
In this part, we study the kernel size in all body stages. 
Inspired by the large local window size in Swin-T~\cite{liu2021swin}, ConvNeXt~\cite{liu2022convnet} boosts the generalization accuracy via increasing the kernel size from $3 \times 3$ to $7 \times 7$.
A large kernel size can extract more semantic information but implicitly increases the attack area during back-propagation. 
It is unclear whether a larger kernel size can bring higher robustness. 
We evaluate kernel size $\in \{ 3, 5, 7 \}$ and find the accuracy grows along with the kernel size (Table \ref{tab:components} row 1, 11 and 12), but the total parameters also increase significantly: kernel $=3$ ($25.56$M), kernel $=5$ ($45.68$M), and kernel $=7$ ($75.86$M). 
Thus, using a large kernel size is a potential candidate to optimize the robustness when scaling up the model. 
We will revisit the design in Sec. \ref{sec:scale_up}. 

\textbf{Guideline 9:} \textit{Purely increasing the kernel size raises the model capacity but improves robustness significantly. Thus, it is a prospective option when scaling up the network.} \par

\smallskip
\noindent \textbf{Dilation.}
Dilated convolution supports the exponential expansion of the receptive field without loss of resolution~\cite{yu2015multi}. 
The operation offers a wider field of view at a similar computational cost. 
However, the results in Table \ref{tab:components} (row 1, 13 and 14) show that a larger dilation factor significantly decreases both natural and \ac{pgd} accuracy after \ac{at}.
Connecting to the previous kernel size section, we hypothesize that a larger receptive field facilitates the attacker. 
We still observe the robustness gain in using a large kernel size because the huge model capacity mitigates the effect, yet the accuracy drops when adjusting dilation since the operation does not change the model capacity.
In Sec. \ref{sec:scale_up}, we also notice the kernel size is not effective in optimizing robustness if all other modifications are considered at the same scale.

\textbf{Guideline 10:} \textit{Increasing dilation factor enlarges the attacking area, which leads to inferior robustness.} \par

\smallskip
\noindent \textbf{Activation.}
We study two factors in the activation layer: the activation function and the number of activation layers in a block. 
For the activation function, we replace \ac{relu}, which is used in ResNet-50, with two smoother functions, \ac{gelu} and \ac{silu}. 
\ac{gelu} alone significantly improves the robustness (Table \ref{tab:components} row 15), which echoes the result in~\cite{bai2021transformers}. 
\ac{silu} further improves the accuracy (Table \ref{tab:components} row 16), which echoes the result in~\cite{xie2020smooth}. 
Recently, Dai \etal~\cite{dai2022parameterizing} added learnable parameters to original non-parametric functions, and proposed the parametric counterparts, \eg, \ac{relu} to \ac{prelu} and \ac{silu} to \ac{psilu} or \ac{pssilu}. 
These parametric functions outperform the non-parametric ones on CIFAR-10~\cite{krizhevsky2009learning}. 
We test these functions on ImageNet and observed \ac{psilu} has the highest robustness among all parametric functions, as shown in Table \ref{tab:components} row 17. 
However, compared to the non-parametric versions, all parametric functions are less robust. 
Since the original paper only tested on the small-scale dataset, we believe such learnable functions are not compatible with the large-scale dataset. 
Next, we reduce the activation layers in each block. 
Neither reducing one nor reducing two activation layers show extra benefits to the robustness.
The more activation layer we reduce, the worse the performance is.

\textbf{Guideline 11:} \textit{Activation function significantly affects robustness.
The non-parametric \ac{silu} provides a competitive improvement.}

\textbf{Guideline 12:} \textit{Reducing activation layers in a residual block severely hurts the robustness.}

\smallskip
\noindent \textbf{\acf{se}.}
Hu \etal~\cite{hu2018squeeze} first introduced the \ac{se} block that explicitly explored inter-dependencies between channels, and adaptively recalibrated channel-wise feature responses.
Inspired by RegNet \cite{radosavovic2020designing}, we place the \ac{se} block between the last two convolutions in each block and set the reduction ratio as $1/4$. 
Compared to ResNet-50, Table \ref{tab:components} row 18 shows that adding \ac{se} significantly improves the robustness.
Directly adding the \ac{se} module slightly increases the model capacity by $2.17$M, but in Sec. \ref{sec:roadmap}, we show that sacrificing the parameters in other components by adopting the \ac{se} module can still improve the robustness, which proves the effectiveness of \ac{se}. 

Since switching activation functions shows significant differences, we also replace \ac{relu} in the \ac{se} block with \ac{silu}, \ac{gelu} and their parametric versions. 
We still observe that non-parametric activation functions are better than their parametric counterparts. 
The \ac{silu} is again the optimal activation for \ac{se} module. 
However, in Sec. \ref{sec:roadmap}, we find that replacing the activation function in activation layers and \ac{se} at the same time causes inferior robustness.  

\textbf{Guideline 13:} \textit{The \ac{se} module significantly contributes to robustness.}

\textbf{Guideline 14:} \textit{The robustness improves if we just replace the activation function in the \ac{se} block. But the modification does not work favorably with switching the activation function in the residual block.}

\smallskip
\noindent \textbf{Normalization.}
Similar to the activation layer, we examine both normalization functions and the number of normalization layers in a block. 
For the normalization function, we switch the original \ac{bn} \cite{ioffe2015batch} in ResNet-50 to \ac{in} \cite{ulyanov2016instance}. 
The training is extremely hard to converge and thus leading to an almost non-robust model (\ac{pgd}10-4: $8.54\%$). 
Then, we attempt to reduce the total normalization layers in a residual block. 
In Table \ref{tab:components}, row 19 to 21 show that reducing the first normalization layer in a residual block optimizes the robustness. 
We keep reducing 2 \acp{bn}, and no further benefits are observed. 

\textbf{Guideline 15:} \textit{Switching \ac{bn} to \ac{in} harms robustness. }

\textbf{Guideline 16:} \textit{Reducing the first \ac{bn} in a residual block benefits robustness.}

\section{Experiments}
\label{sec:experiments}

In this section, we provide a roadmap that outlines the path we take to construct the \ac{ra} using the guidelines in Sec. \ref{sec:robust_arch_design}. 
Our roadmap combines architecture components such that for each combination we only keep components that increase robustness. 
Then, we scale up the resulting model and proposed a family of \ac{ra} models.
Finally, we compare \ac{ra} with other \ac{sota} architectures. 
See Appendix \ref{sec:appendix-exp-setting} for the full experimental setup.
We also ablate Fast-\ac{at} and Standard-\ac{at} in Appendix \ref{sec:appendix-ablation}. 

\begin{table}
\small
  \centering
  \caption{The roadmap outlines the path we take to cumulatively improve the robustness and construct \ac{ra}-S (${\sim}26$M), \ac{ra}-M (${\sim}46$M), and \ac{ra}-L (${\sim}104$M) based on our guidelines. \ac{pgd}10-2 and \ac{pgd}10-8 show a similar trend of accuracy improvement as \ac{pgd}10-4, and detailed results are shown in Appendix \ref{sec:appendix-roadmap}.}
  \begin{tabular}{@{}l@{\hspace{0.6mm}}|lr@{\hspace{2mm}}r@{}}
  \toprule
  & Configurations &  Natural & \ac{pgd}10-4\\
  \midrule
  \multicolumn{4}{c}{\textit{Small}: ResNet-50 $\rightarrow$ \ac{ra}-S ($\mathcal{S}_{7}$)} \\
  \midrule
  $\mathcal{S}_0$ & ResNet-50 & 56.09\% & 30.43\% \\ 
  $\mathcal{S}_1$ & $\mathcal{S}_0$ + $D\mh 5\mh 8\mh 13\mh 1$ & 57.35\% & 33.33\% \\
  $\mathcal{S}_{2a}$ & $\mathcal{S}_1$ + $g=2, e=2, b=0.25$ & 57.98\% & 33.94\% \\
  $\mathcal{S}_{2b}$ & $\mathcal{S}_1$ + $g=1, e=1.5, b=0.25$ & 57.52\% & 32.83\% \\ 
  \multirow{2}{*}{$\mathcal{S}_{3}$} & $\mathcal{S}_{2a}$ + Stem width 96 & \multirow{2}{*}{57.82\%} & \multirow{2}{*}{34.86\%} \\
  & + Move down ($\downarrow$) downsampling & & \\
  $\mathcal{S}_{4}$ & $\mathcal{S}_{3}$ + \ac{se} (\ac{relu})  & 60.57\% & 36.61\% \\
  $\mathcal{S}_{5}$ & $\mathcal{S}_{4}$ + Act. \ac{silu} & 62.04\% & 39.48\% \\
  $\mathcal{S}_{6}$ & $\mathcal{S}_{5}$ + \ac{se} (\ac{silu}) & 60.32\% & 38.24\% \\
  $\mathcal{S}_{7}$ & $\mathcal{S}_{5}$ + Norm$\mh$0$\mh$BN$\mh$BN & 62.27\% & 39.88\% \\
  \midrule
  \multicolumn{4}{c}{\textit{Medium}: \ac{ra}-S ($\mathcal{S}_{7}$) $\rightarrow$ \ac{ra}-M ($\mathcal{M}_{2}$)} \\
  \midrule
  $\mathcal{M}_{1}$ & $\mathcal{S}_{7}$ + Kernel size 5 & 63.82\% & 41.00\% \\
  $\mathcal{M}_{2}$ & $\mathcal{S}_{7}$ + $D\mh 7\mh 11\mh 18\mh 1$ & 64.40\% & 42.06\% \\
  $\mathcal{M}_{3}$ & $\mathcal{S}_{7}$ + $W\mh 384\mh 760\mh 1504\mh 2944$ & 63.52\% & 41.43\% \\
  \midrule
  \multicolumn{4}{c}{\textit{Large}: \ac{ra}-M ($\mathcal{M}_{2}$) $\rightarrow$ \ac{ra}-L ($\mathcal{L}_{2}$)} \\
  \midrule
  $\mathcal{L}_{1}$ & $\mathcal{M}_{2}$ + Kernel size 7 & 64.08\% & 40.70\% \\
  $\mathcal{L}_{2}$ & $\mathcal{M}_{2}$ + $W\mh 512\mh 1024\mh 2016\mh 4032$ & 66.08\% & 43.81\% \\
  $\mathcal{L}_{3}$ & $\mathcal{M}_{2}$ + $D\mh 8\mh 13\mh 21\mh 2$ & 64.91\% & 43.09\% \\
  $\mathcal{L}_{4}$ & $\mathcal{M}_{2}$ + $D\mh 10\mh 16\mh 26\mh 2$ & 65.28\% & 42.85\% \\
  \bottomrule
  \end{tabular}
  \label{tab:robarch}
\end{table}

\subsection{A Roadmap from ResNet-50 to \ac{ra}-S}
\label{sec:roadmap}

In this section, we cumulatively construct \ac{ra}-S from ResNet-50 based on the proposed guidelines.
Table \ref{tab:robarch} (upper) presents the procedures and results at each step of network modification. 
We start with network depth and width. 
Combining guideline 2 and guideline 5, model $\mathcal{S}_1$ selects the optimal depth configuration $D \mh 5 \mh 8 \mh 13 \mh 1$.
For width, we test the two optimal width configurations in guideline 4 and select $g = 2, e = 2$ ($\mathcal{S}_{2a}$). 
For the stem stage, model $\mathcal{S}_3$ increases the width to $96$ and replace the max-pooling in the stem stage with a downsampling shortcut in the first stage according to guidelines 6 and 7. 
Then, we optimize the block settings in each stage.
Guideline 13 suggests inserting a \ac{se} block between the last 2 convolutions.
To accommodate the extra parameters in the modification, we reduce the width in all stages and build model $\mathcal{S}_4$. 
Next, $\mathcal{S}_5$ substitutes \ac{silu} for \ac{relu} in all 3 activation layers. 
However, we find that continuing to replace the activation function in the \ac{se} block lowers the robustness. 
Thus, we discard the modification, reduce the first \ac{bn} layer, and construct $\mathcal{S}_7$. 
The resulting model is named \ac{ra}-S. 
The guidelines are verified by the consistent increase in robustness along the network construction process. 
The total model capacity is comparable to ResNet-50, but both natural and \ac{pgd}-4 accuracies have increased by $6.18$ and $9.45$ percentage points, respectively. 

\begin{figure}[t]
\centering
    \includegraphics[width=0.9\linewidth]{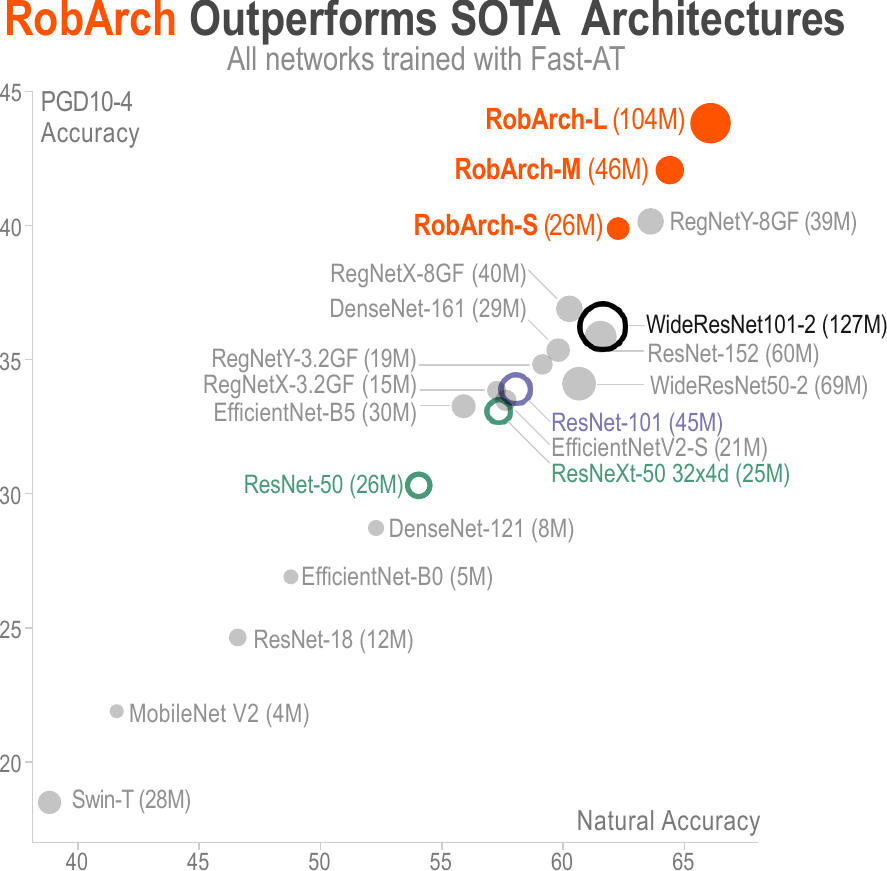}
    \caption{
    Our \textbf{\ac{ra}} model family outperforms \ac{sota} architectures under the same Fast-\ac{at} training method. 
    With a similar model capacity, \textbf{\ac{ra}-S} outperforms 
    \mcolor{mygreen}{\textbf{ResNet-50}}~\cite{he2016deep} and 
    \mcolor{mygreen}{\textbf{ResNeXt-50 32$\times$4d}}~\cite{xie2017aggregated} by 9.45 and 6.80 percentage points, respectively. \textbf{\ac{ra}-M} outperforms \mcolor{mypurple}{\textbf{ResNet-101}} by 8.16 percentage points.
    Compared to the models with larger parameters, \textbf{\ac{ra}-S} is even more robust than \mcolor{mygray}{\textbf{WideResNet101-2}} despite having $4.85\times$ fewer parameters (highlighted in black). 
    Appendix \ref{sec:appendix-roadmap} shows detailed results, including other \ac{pgd} attack budgets.
    }
\label{fig:fat-sota}
\end{figure}

\begin{table}[t]
\small
\centering
\caption{
Our \ac{ra} model outperforms \textit{ConvNets} and \textit{Transformers} with similar total parameters against $\ell_\infty = 4 / 255$ \ac{aa}. 
Using the same training configurations as Salman \etal \cite{salman2020adversarially}, our model outperforms both ResNet-50 and WideResNet50-2. 
Every \ac{ra} model outperforms its \ac{xcit} counterpart at a similar capacity. 
Appendix \ref{sec:appendix-aa} shows the detailed results including \ac{pgd}100 for $\epsilon \in \{2, 4, 8 \}$.}
\begin{tabular}{@{}l@{}rrr@{}}
\toprule
    Architecture & \#Param & \acl{aa} & Natural\\
\midrule
    ResNet-18 \cite{salman2020adversarially}   & 12M & 25.32\%          & 52.49\% \\
    PoolFormer-M12 \cite{debenedetti2022light} & 22M & 34.72\%          & 66.16\% \\
    DeiT-S \cite{bai2021transformers}          & 22M & 35.50\%          & 66.50\% \\
    DeiT-S+DiffPure \cite{nie2022diffusion}    & 22M & 43.18\%          & 73.63\% \\
    ResNet-50 \cite{salman2020adversarially}   & 26M & 34.96\%          & 63.87\% \\
    ResNet-50+DiffPure \cite{nie2022diffusion} & 26M & 40.93\%          & 67.79\% \\
    ResNet-50+GELU \cite{bai2021transformers}   & 26M & 35.51\%          & 67.38\% \\
    XCiT-S12 \cite{debenedetti2022light}       & 26M & 41.78\%          & 72.34\% \\
    \textbf{\ac{ra}-S}                         & 26M & \textbf{44.14\%} & 70.17\% \\
\midrule
    XCiT-M12 \cite{debenedetti2022light} & 46M & 45.24\%          & 74.04\% \\
    \textbf{\ac{ra}-M}                   & 46M & \textbf{46.26\%} & 71.88\% \\
\midrule
    WideResNet50-2 \cite{salman2020adversarially} & 69M & 38.14\% & 68.41\% \\
    WideResNet50-2                                & \multirow{2}{*}{69M} & \multirow{2}{*}{44.39\%} & \multirow{2}{*}{71.16\%} \\
    +DiffPure \cite{nie2022diffusion}             &  &  &  \\
    Swin-B \cite{mo2022adversarial}               & 88M  & 38.61\%          & 74.36\% \\
    XCiT-L12 \cite{debenedetti2022light}          & 104M & 47.60\%          & 73.76\% \\
    \textbf{\ac{ra}-L}                            & 104M & \textbf{48.94\%} & 73.44\% \\
\bottomrule
\end{tabular}
\label{tab:sota_paper}
\end{table}

\subsection{Scaling Up: The \ac{ra} Family}
\label{sec:scale_up}
We extend our investigation to optimize the robustness when scaling up the parameter budget. 
The budgets align with the \ac{xcit}~\cite{ali2021xcit} family since it is the current \ac{sota} on the RobustBench ImageNet leaderboard~\cite{croce2020robustbench}. 
Guideline 9 suggests increasing kernel size as a potential improvement when scaling up the model. 
Increasing total depth and width are another 2 promising directions~\cite{xie2019intriguing, huang2021exploring}. 
For the medium-sized budget (${\sim}46$M), model $\mathcal{M}_{1}$ enlarges the kernel size from 3 to 5, model $\mathcal{M}_{2}$ proportionally deepens the network by a factor of 1.4, and model $\mathcal{M}_{3}$ widens the channels while keeping the depth same as \ac{ra}-S.
The training results of $\mathcal{M}_{1}$, $\mathcal{M}_{2}$ and $\mathcal{M}_{3}$ are shown in Table \ref{tab:robarch} (middle). 
In general, all three models are more robust than \ac{ra}-S. 
But in terms of accuracy, increasing depth ($\mathcal{M}_{2}$) $>$ increasing width ($\mathcal{M}_{3}$) $>$ increasing kernel size ($\mathcal{M}_{1}$). 
Therefore, we set $\mathcal{M}_{2}$ as \ac{ra}-M. 

For the large-sized budget (${\sim}104$M), model $\mathcal{L}_{1}$ increases the kernel size from 3 to 7, but leads to a drop in robustness, as shown in Table \ref{tab:robarch} (bottom). 
\ac{ra}-M increases the depth of $\mathcal{S}_{7}$, and according to the depth-width trade-off in Fig. \ref{fig:depth-width}, consistently increasing the depth can lead to unstable training. 
Therefore, model $\mathcal{L}_{2}$ increases the width in \ac{ra}-M, and the robustness rises by a large margin. 
We further deepen $\mathcal{L}_{2}$ to explore whether guideline 5 holds true when scaling up the model budget. 
$\mathcal{L}_{3}$ and $\mathcal{L}_{4}$ increase the depth by $1.6\times$ and $2\times$ and reduce the width to fit the total parameters. 
The results in Table \ref{tab:robarch} (bottom) show a decline in accuracy along with an increase in depth. 
The phenomenon extends guideline 5 that the depth-width relationship also applies to scaling up the models. 
Finally, we set $\mathcal{L}_{2}$ as \ac{ra}-L based on the above discussions, and provide the following guidelines: 

\textbf{Guideline 17:} \textit{When scaling up the model, increasing the kernel size, depth, and width all contribute to the robustness. But proportionally increasing the optimal depth configuration is most effective.}

\textbf{Guideline 18:} \textit{There exists a saturation point for purely increasing the depth to fill the parameter budget. We should enlarge channel widths when such a degradation happens.}

\subsection{Results}
\label{sec:results}

In Fig.~\ref{fig:fat-sota}, we compare \ac{ra} with a series of \ac{sota} architectures. 
All architectures are trained with Fast-\ac{at} for a fair comparison, and we discover a similar trend for \ac{pgd}10-2, \ac{pgd}10-4, and \ac{pgd}10-8. 
Below we provide a few observations based on \ac{pgd}10-4 accuracy:
\smallskip
\begin{enumerate}[label=\arabic*), leftmargin=*, topsep=0pt, wide=0pt]
\itemsep-0.2em 
    \item Under a similar model capacity, \textbf{\ac{ra}-S} outperforms ResNet-50~\cite{he2016deep} and ResNeXt-50 32$\times$4d~\cite{xie2017aggregated} by 9.45 and 6.80 percentage points, respectively. \textbf{\ac{ra}-M} outperforms ResNet-101 by 8.16 percentage points.
    \item Compared to the models with larger parameters, \textbf{\ac{ra}-S} is even more robust than WideResNet101-2 despite having $4.85\times$ fewer parameters.
    \item Increasing the total parameters in general leads to higher robustness, and the natural accuracy is positively correlated with the adversarial accuracy after \ac{at}. Lightweight models, \eg, MobileNet V2 and SqueezeNet-1.1, are among the least robust. The accuracies of \textbf{\acp{ra}} consistently grow when scaling up the model sizes.
    \item Transformers, \eg, Swin-T~\cite{liu2021swin}, and Transformer-based architectures, \eg, ConvNeXt-T~\cite{liu2022convnet}, are non-robust using Fast-\ac{at}. The phenomenon can be attributed to the differences in optimizers and learning rates, where most Transformer-related architectures use AdamW~\cite{loshchilov2017decoupled} and tiny learning rates.
\end{enumerate}

As introduced in Sec. \ref{sec:training}, we then train all \acp{ra} using Standard-\ac{at}. 
All three \acp{ra} outperform their \ac{xcit}~\cite{ali2021xcit} counterparts (Table \ref{tab:sota_paper}). 
Using the same training configurations as Salman \etal~\cite{salman2020adversarially}, \textbf{\ac{ra}-S} surpasses ResNet-50 \ac{aa} accuracy by $9.18$ percentage points, and is even more robust than WideResNet50-2 with 2.6$\times$ fewer parameters. 
The robustness continues to improve when scaling up the model, and \textbf{\ac{ra}-L} achieves the new \ac{sota} \ac{aa}~\cite{croce2020reliable} accuracy on RobustBench. 
\ac{ra}'s performance advantage also extrapolates to the \ac{pgd} attack. 
Overall, the proposed \acp{ra} outperform both \textit{ConvNets} and \textit{Transformers} with similar total parameters. 

\section{Related Work}
\label{sec:related_work}

A huge number of \ac{at} variants have been proposed, \eg, 
TRADES~\cite{zhang2019theoretically},
AWP~\cite{wu2020adversarial},
ADT~\cite{dong2020adversarial},
DART~\cite{wang2021convergence},
MART~\cite{wang2019improving},
CAS~\cite{bai2021improving},
Max-Margin AT \cite{ding2018mma}, \etc.
For the robust \ac{dnn} research, only a few studies explored how architectures affect robustness~\cite{tang2021robustart, madry2018towards, devaguptapu2021adversarial, su2018robustness}, \eg, depths~\cite{xie2019intriguing}, widths~\cite{wu2021wider} and activation functions~\cite{xie2020smooth, dai2022parameterizing}. 
However, the total model capacity is unconstrained along with the architecture modifications.
Besides, simply combining multiple individual optimal architectures does not transfer to a better model, \eg, Huang \etal~\cite{huang2021exploring} studied depths and widths, and found the combination of the optimal depth and width ratios is less robust than just using the optimal width ratio.

\section{Conclusion}
\label{sec:conclusion}

In this work, we present the first large-scale systematic study on the robustness of architecture components under fixed parameter budgets.
Through our investigation, we distill 18 actionable robust network design guidelines that empower model developers to gain deep insights.
Our \ac{ra} models instantiate the guidelines to build a family of top-performing models across parameter capacities against strong adversarial attacks.

\clearpage
{\small
\bibliographystyle{ieee_fullname}
\bibliography{egbib}
}

\clearpage
\appendix
\input{supplementary.tex}

\end{document}

%% file: supplementary.tex
\section{Network Configurations}
\label{sec:appendix-network}

\subsection{Overview of ResNet-style ConvNets}
\label{sec:appendix-resnet50}

\begin{figure}[htbp]
\centering
    \includegraphics[width=0.9\linewidth]{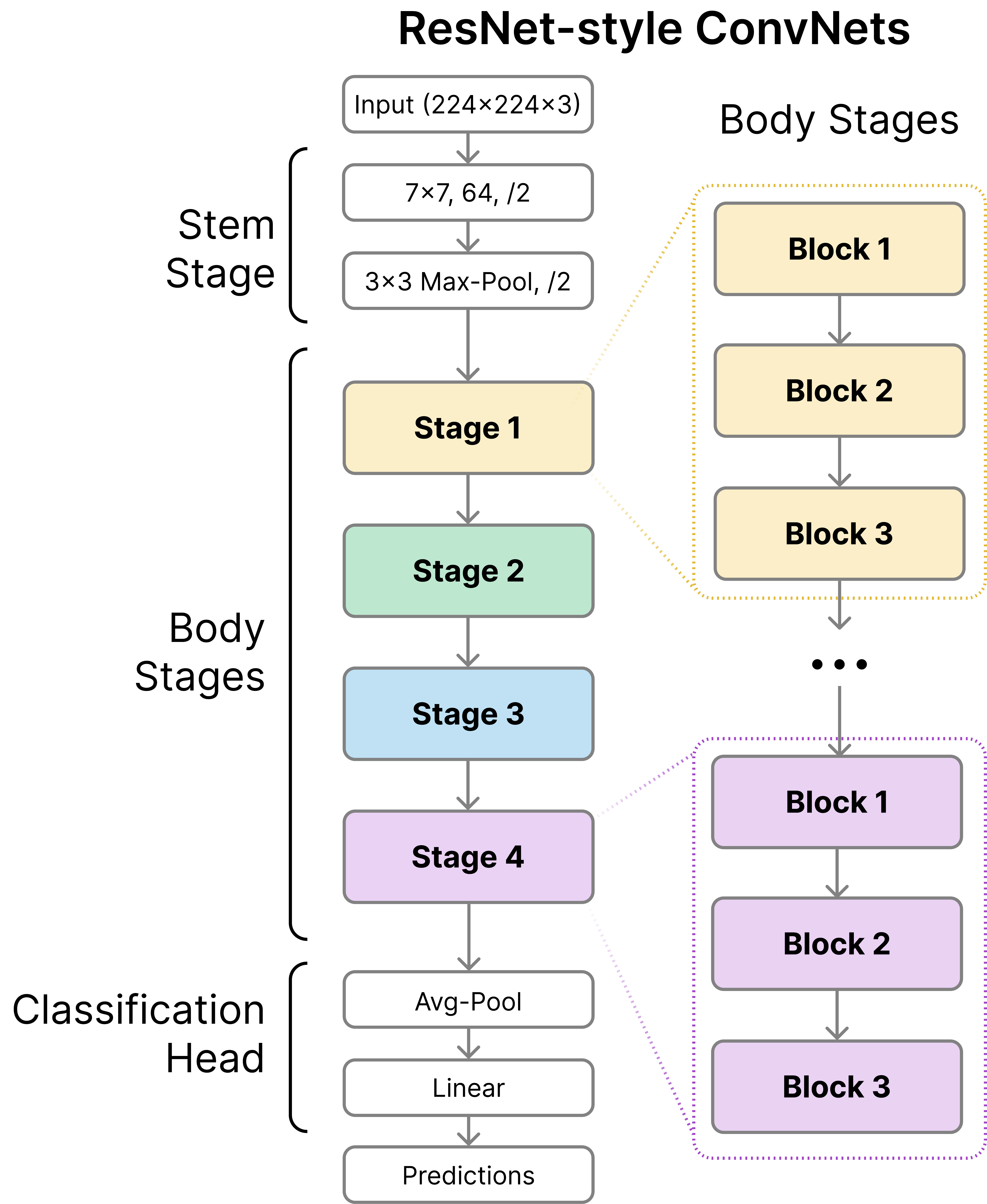}
    \caption{
    An overview of ResNet-style ConvNet design, which includes a stem stage, several body stages, and a classification head. 
    }
\label{fig:resnet50-arch}
\end{figure}

A standard ResNet-style ConvNet includes a stem stage, several body stages, and a classification head, as shown in Fig. \ref{fig:resnet50-arch}. 
A typical body stage consists of multiple residual blocks, where each of them has a shortcut connection that skips other layers and feeds the output of the previous layer to the current output of the block \cite{he2016deep}. 
The stem stage proceeds the input image through a convolution layer and a max-pooling that downsample the resolution by 4 in total. 
The final classification head passes the extracted features from body stages through an average pooling and a linear layer that outputs the predictions. 
Table \ref{tab:full-resnet50} lists ResNet-50 configurations written in notations defined in the paper.

\subsection{RobArch Architecture}
\label{sec:appendix-robarch}

\begin{figure}[htbp]
\centering
    \includegraphics[width=0.9\linewidth]{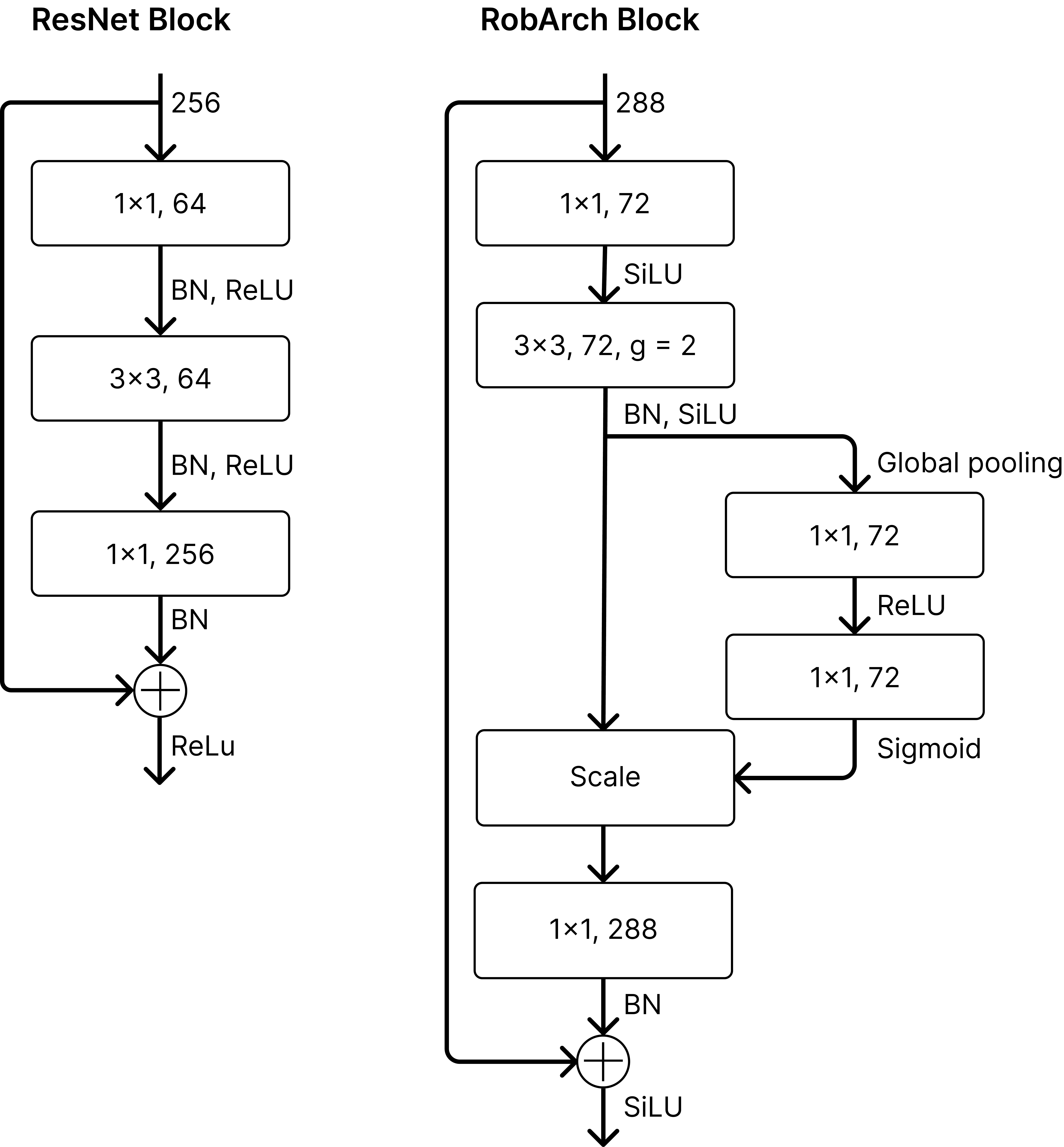}
    \caption{
    Block designs for a ResNet and a \ac{ra}. 
    For simplicity, ``$1 \times 1$, 64'' means pointwise convolution with 64-layer output channels. 
    ``$g = 2$'' means 2 group convolution groups, and the default group is 1. 
    }
\label{fig:block-design}
\end{figure}

The \ac{ra} follows the ResNet-style ConvNet design. 
We display block designs for ResNet-50 and \ac{ra}-S in Fig. \ref{fig:block-design}. 
Following the RegNet design~\cite{radosavovic2020designing}, we add the \ac{se} block
after the $3 \times 3$ convolution layer in each block.  
The \ac{se} reduction ratio is $0.25$. 

\begin{table}
\small
\centering
\caption{ResNet-50 configurations written in notations defined in the paper. The left column lists architecture components, and the right column shows notations. ResNet-50 does not have \ac{se} block, so the configuration is ``N/A''. For activation, \ac{relu}$\mh$\ac{relu}$\mh$\ac{relu} represents the three activation layers in a residual block. The same also applies to normalization. }
\setlength{\tabcolsep}{4mm}
\begin{tabular}{ll}
\toprule
    & Notation  \\
\midrule
\rowcolor{mylightgray}
    Depth & $D\mh 3\mh 4\mh 6\mh 3$ \\
    Width & $W\mh256\mh512\mh1024\mh2048$ \\
          & $G\mh1\mh1\mh1\mh1$ \\
          & $BM\mh0.25\mh0.25\mh0.25\mh0.25$ \\
\rowcolor{mylightgray}
    Stem stage & Stem width 64 \\
\rowcolor{mylightgray}
          & Stem kernel 7 \\
\rowcolor{mylightgray}
          & Downsample factor 4 \\
    Dense connection & Dense ratio 1 \\
\rowcolor{mylightgray}
    Kernel size & Kernel size 3 \\
    Dilation & Dilation 1 \\
\rowcolor{mylightgray}
    Activation & Act. \ac{relu} \\
\rowcolor{mylightgray}
          & \ac{relu}$\mh$\ac{relu}$\mh$\ac{relu} \\
    \ac{se} & N/A \\
\rowcolor{mylightgray}
    Normalization & Norm. \ac{bn} \\
\rowcolor{mylightgray}
          & \ac{bn}$\mh$\ac{bn}$\mh$\ac{bn} \\
\bottomrule
\end{tabular}
\label{tab:full-resnet50}
\end{table}

\section{Experimental Settings}
\label{sec:appendix-exp-setting}

We use Fast-\ac{at} as a rapid indicator while exploring different architecture components and building the \ac{ra} family. 
We follow the same 3-phase training as proposed in the Fast-\ac{at} paper \cite{wong2020fast}.
Fast-\ac{at} sets training $\epsilon = 1.25 \times$ test $\epsilon$ and finds catastrophic overfitting happens when training $\epsilon$ goes beyond $10$. 
Therefore, we set training $\epsilon \in \{ 2.5, 5.0, 7.5\}$ corresponding to test $\epsilon \in \{2, 4, 6 \}$ and show the results in Table \ref{tab:full-exp}.
Larger training $\epsilon$ exhibits higher robustness against strong attacks at the cost of lowering the accuracies of natural and weak attacks. 
We select training $\epsilon = 5.0$ for its balanced performance on natural and various attack budgets. 
We use Standard-\ac{at} to robustify all members in the \ac{ra} family, and follow the same training configurations as Salman \etal~\cite{salman2020adversarially}. 

Our \acp{ra} are evaluated against the two strongest adversarial attacks, \ac{pgd}~\cite{madry2018towards} and \ac{aa}~\cite{croce2020reliable}.
All \ac{pgd} attacks are tested on the full ImageNet validation set. 
\ac{aa} is an ensemble of four different parameter-free attacks, three
white- and one black-box.
We use the same 5000 ImageNet validation subset provided by the RobustBench~\cite{croce2020robustbench} for \ac{aa} comparison. 

\begin{table}
\small
\centering
\caption{Determine training $\epsilon$ for Fast-\ac{at} using ResNet-50. Training $\epsilon = 2.5$ shows the highest natural and \ac{pgd}10-2 accuracies, while training $\epsilon = 7.5$ shows the highest \ac{pgd}10-8 accuracy. Overall, training $\epsilon = 5.0$ is selected for all Fast-\ac{at} experiments since it exhibits a balanced performance on all natural and attack budgets.}
\begin{tabular}{rrrrr}
\toprule
    Training $\epsilon$ & Natural & PGD10$\mh$2 & PGD10$\mh$4 & PGD10$\mh$8 \\
\midrule
    2.5 & 60.04\% & 43.06\% & 25.34\% &  6.49\% \\ 
    5.0 & 56.09\% & 42.66\% & 30.43\% & 12.61\% \\
    7.5 & 49.80\% & 36.86\% & 26.95\% & 13.87\% \\ 
\bottomrule
\end{tabular}
\label{tab:full-exp}
\end{table}

\section{Ablations on Adversarial Training}
\label{sec:appendix-ablation}

We ablate Fast-\ac{at} and Standard-\ac{at} for two purposes:
1) verify the robustness order is consistent under two different \ac{at} methods,
2) compute whether the two approaches exhibit comparable robustness increases when subjected to the same ablation.

Since Standard-\ac{at} incurs longer training time, we randomly select one small budget model $\mathcal{S}_{4}$ and show the results in Table \ref{tab:full-ablation}. 
For natural, \ac{pgd}10-4 and \ac{aa} runs, $\mathcal{S}_{4}$ outperforms ResNet-50 but is inferior to \ac{ra}-S, which demonstrates the robustness order is consistent under Fast-\ac{at} and Standard-\ac{at}. 
Then, we compute the robustness gain using \ac{pgd}10-4 as an example. 
From ResNet-50, $\mathcal{S}_{4}$ to \ac{ra}-S, accuracy increases by $6.18$ and $3.27$ percentage points under Fast-\ac{at}, and increases by $6.01$ and $2.52$ percentage points under Standard-\ac{at}. 
Both training methods show comparable robustness increases on the same architecture against the same attack. 
The observation also extrapolates to natural and \ac{aa} accuracies. 
As expected, Standard-\ac{at} displays higher robustness than Fast-\ac{at}.
Hence, we conclude that Fast-\ac{at} serves as a good indicator when exploring different architecture components and building the \ac{ra} family. 
Standard-\ac{at} can fully robustify all members in the \ac{ra} family after finalizing the architectures.

\begin{table}
\small
\centering
\caption{Ablations on Fast-\ac{at} and Standard-\ac{at}. We randomly select one small budget model $\mathcal{S}_{4}$ from the roadmap and train it with both methods. The results show that the robustness order is consistent under two different \ac{at} methods, and the scales of robustness increment are also comparable.}
\begin{tabular}{@{}l@{\hspace{2mm}}r@{\hspace{1.5mm}}r@{\hspace{2.5mm}}r@{\hspace{1.5mm}}r@{\hspace{1.5mm}}r@{}}
\toprule
    \multirow{2}{*}{Model} & \multicolumn{2}{c}{Fast-\ac{at}} & \multicolumn{3}{c}{Standard-\ac{at}} \\
    & Natural & PGD10$\mh$4 & Natural & PGD10$\mh$4 & \ac{aa} \\
\midrule
    $\mathcal{S}_{0}$ (ResNet-50) & 56.09\% & 30.43\% & 63.87\% & 39.66\% & 34.96\% \\
    $\mathcal{S}_{4}$             & 60.57\% & 36.61\% & 68.88\% & 45.67\% & 41.44\% \\
    $\mathcal{S}_{7}$ (\ac{ra}-S) & 62.27\% & 39.88\% & 70.17\% & 48.19\% & 44.14\% \\
\bottomrule
\end{tabular}
\label{tab:full-ablation}
\end{table}

\begin{table}
\small
\centering
\caption{Our \ac{ra} model family outperforms \ac{sota} architectures under the same Fast-\ac{at} training method. The results are consistent across natural and different attack budgets. We highlight all three \acp{ra} for easy comparisons. }
\begin{tabular}{>{\kern-\tabcolsep}l@{\hspace{-3.9mm}}r@{\hspace{1.8mm}}r@{\hspace{1.9mm}}r@{\hspace{1.9mm}}r@{\hspace{1.9mm}}r<{\kern-\tabcolsep}}
\toprule
     Architecture & \#Param & Natural & PGD10$\mh$2 & PGD10$\mh$4 & PGD10$\mh$8 \\
\midrule
    SqueezeNet 1.1            &  1 M & 0.10 \% & 0.10 \% & 0.10 \% & 0.10 \% \\
    MobileNet V2              &  4 M & 41.60\% & 31.23\% & 21.89\% & 8.94 \% \\
    EfficientNet-B0           &  5 M & 48.78\% & 37.74\% & 26.90\% & 10.92\% \\
    ShuffleNet V2 2.0$\times$ &  7 M & 49.99\% & 0.01 \% & 0.01 \% & 0.02 \% \\
    DenseNet-121              &  8 M & 52.29\% & 40.06\% & 28.72\% & 12.23\% \\
    ResNet-18                 & 12 M & 46.59\% & 35.05\% & 24.64\% & 9.95 \% \\
    RegNetX-3.2GF             & 15 M & 57.26\% & 45.74\% & 33.85\% & 15.37\% \\
    RegNetY-3.2GF             & 19 M & 59.15\% & 47.09\% & 34.82\% & 15.51\% \\
    EfficientNetV2-S          & 21 M & 57.64\% & 45.89\% & 33.48\% & 14.03\% \\
    ResNeXt-50                & \multirow{2}{*}{25M} & \multirow{2}{*}{57.33\%} & \multirow{2}{*}{45.46\%} & \multirow{2}{*}{33.08\%} & \multirow{2}{*}{14.45\%} \\
    32$\times$4d              &      &         &         &         &         \\
    ResNet-50                 & 26 M & 56.09\% & 42.66\% & 30.43\% & 12.61\% \\
\rowcolor{myorange}
    \ac{ra}-S                 & 26 M & 62.27\% & 51.67\% & 39.88\% & 18.99\% \\
    Swin-T                    & 28 M & 38.83\% & 28.08\% & 18.49\% & 6.20 \% \\
    ConvNeXt-T                & 29 M & 21.35\% & 15.39\% & 10.51\% & 4.07 \% \\
    DenseNet-161              & 29 M & 59.80\% & 47.60\% & 35.35\% & 15.77\% \\
    EfficientNet-B5           & 30 M & 55.90\% & 44.80\% & 33.26\% & 14.53\% \\
    RegNetY-8GF               & 39 M & 63.61\% & 52.26\% & 40.15\% & 19.21\% \\
    RegNetX-8GF               & 40 M & 60.26\% & 48.98\% & 36.89\% & 17.22\% \\
    ResNet-101                & 45 M & 58.04\% & 45.72\% & 33.90\% & 15.93\% \\
\rowcolor{myorange}
    \ac{ra}-M                 & 46 M & 64.40\% & 53.97\% & 42.06\% & 20.98\% \\
    ResNet-152                & 60 M & 61.55\% & 48.50\% & 35.85\% & 15.87\% \\
    WideResNet50-2            & 69 M & 60.66\% & 46.99\% & 34.10\% & 15.37\% \\
\rowcolor{myorange}
    \ac{ra}-L                 & 104M & 66.08\% & 55.52\% & 43.81\% & 22.50\% \\
    WideResNet101-2           & 127M & 61.63\% & 49.10\% & 36.23\% & 16.14\% \\
\bottomrule
\end{tabular}
\label{tab:full-fat}
\end{table}

\section{Robust Architecture Design Results}
\label{sec:appendix-arch-components}

This section presents the detailed results for all architecture components, using five tables.   
In each table, we use a bold font to highlight the results that have been presented in the paper, and in the caption, we describe the additional information that we are introducing here. 
Table \ref{tab:full-depth-stage} is depth-only, Table \ref{tab:full-width} is width-only, Table \ref{tab:full-depth-width} is depth-width combination, Table \ref{tab:full-stage} includes all stage-level designs, and Table \ref{tab:full-block} includes all block-level designs. 
For each component, its table includes architecture configurations, total parameters, natural, \ac{pgd}10-2, \ac{pgd}10-4, and \ac{pgd}10-8 accuracies.

\begin{table}
\small
\centering
\caption{\ac{pgd}10 robustness of depth. Bold font means the results have been presented in the paper. All configurations are trained with Fast-\ac{at} and evaluated on full ImageNet validation set. ResNet-50 serves as the baseline. We presented $D \mh 5  \mh 8  \mh 13 \mh 1$ in the main paper, and provide results for all 3-, 4-, 5- and 6-stage networks here. }
\begin{tabular}{@{}l@{\hspace{1.5mm}}r@{\hspace{1.5mm}}r@{\hspace{1.5mm}}r@{\hspace{1.5mm}}r@{\hspace{1.5mm}}r@{}}
\toprule
    Config & \#Param & Natural & PGD10$\mh$2 & PGD10$\mh$4 & PGD10$\mh$8 \\
\midrule
    ResNet-50 & 25.56M & 56.09\% & 42.66\% & 30.43\% & 12.61\% \\
\midrule
  \multicolumn{6}{c}{\textit{3-stage Network}}\\
\midrule
    $D \mh 16 \mh 16 \mh 16$ & 25.02M & 57.15\% & 41.35\% & 29.57\% & 14.37\% \\
    $D \mh 10 \mh 18 \mh 16$ & 25.15M & 56.47\% & 43.32\% & 31.52\% & 14.57\% \\
    $D \mh 3  \mh 22 \mh 16$ & 25.78M & 56.77\% & 44.99\% & 33.24\% & 15.14\% \\
    $D \mh 16 \mh 25 \mh 14$ & 25.30M & 56.69\% & 44.53\% & 32.63\% & 14.39\% \\
    $D \mh 2  \mh 16 \mh 18$ & 26.26M & 57.31\% & 44.97\% & 32.72\% & 14.00\% \\
    $D \mh 3  \mh 29 \mh 14$ & 25.51M & 57.27\% & 44.74\% & 33.02\% & 14.88\% \\
    $D \mh 3  \mh 4  \mh 20$ & 25.21M & 57.69\% & 45.27\% & 32.95\% & 14.46\% \\
    $D \mh 8  \mh 2  \mh 20$ & 25.00M & 57.12\% & 44.32\% & 31.90\% & 13.50\% \\
\midrule
  \multicolumn{6}{c}{\textit{4-stage Network}}\\
\midrule
    $D \mh 1  \mh 5  \mh 6  \mh 3$  & 25.70M & 55.98\% & 43.54\% & 31.46\% & 13.54\% \\
    $D \mh 5  \mh 2  \mh 6  \mh 3$  & 25.14M & 52.93\% & 40.41\% & 29.14\% & 12.60\% \\
    $D \mh 1  \mh 4  \mh 7  \mh 3$  & 26.53M & 56.60\% & 43.62\% & 31.51\% & 13.76\% \\
    $D \mh 6  \mh 4  \mh 4  \mh 3$  & 23.53M & 54.19\% & 42.11\% & 30.40\% & 13.30\% \\
    $D \mh 3  \mh 5  \mh 2  \mh 4$  & 25.83M & 53.98\% & 41.44\% & 30.08\% & 13.13\% \\
    $D \mh 4  \mh 3  \mh 10 \mh 2$  & 25.35M & 55.62\% & 43.15\% & 31.32\% & 14.03\% \\
    $D \mh 2  \mh 7  \mh 13 \mh 1$  & 25.22M & 57.19\% & 44.16\% & 31.91\% & 13.89\% \\
    $D \mh 2  \mh 9  \mh 13 \mh 1$  & 25.78M & 57.89\% & 45.08\% & 32.84\% & 14.63\% \\
    $D \mh 2  \mh 13 \mh 8  \mh 2$  & 25.78M & 55.86\% & 42.91\% & 30.96\% & 13.40\% \\
    $D \mh 1  \mh 1  \mh 15 \mh 1$  & 25.71M & 55.74\% & 43.41\% & 31.45\% & 13.51\% \\
    $D \mh 2  \mh 5  \mh 14 \mh 1$  & 25.78M & 56.49\% & 44.13\% & 32.58\% & 14.73\% \\
    $D \mh 5  \mh 8  \mh 13 \mh 1$  & 25.71M & \textbf{57.35\%} & 44.83\% & \textbf{33.33\%} & 15.46\% \\
    $D \mh 2  \mh 12 \mh 12 \mh 1$  & 25.51M & 55.89\% & 43.39\% & 31.45\% & 13.56\% \\
    $D \mh 4  \mh 8  \mh 1  \mh 4$  & 25.62M & 54.84\% & 42.44\% & 30.23\% & 12.86\% \\
    $D \mh 1  \mh 4  \mh 2  \mh 4$  & 25.41M & 52.46\% & 40.25\% & 28.80\% & 12.22\% \\
    $D \mh 2  \mh 1  \mh 3  \mh 4$  & 25.76M & 53.23\% & 41.50\% & 29.76\% & 12.51\% \\
    $D \mh 3  \mh 24 \mh 5  \mh 2$  & 25.58M & 57.41\% & 44.66\% & 32.65\% & 14.42\% \\
    $D \mh 2  \mh 8  \mh 5  \mh 3$  & 25.49M & 56.43\% & 43.65\% & 31.70\% & 13.62\% \\
    $D \mh 6  \mh 4  \mh 2  \mh 4$  & 25.76M & 53.48\% & 42.04\% & 31.07\% & 13.70\% \\
    $D \mh 10 \mh 6  \mh 5  \mh 3$  & 25.49M & 57.17\% & 43.65\% & 31.45\% & 13.25\% \\
    $D \mh 10 \mh 2  \mh 2  \mh 4$  & 25.48M & 53.03\% & 41.01\% & 30.32\% & 13.45\% \\
    $D \mh 1  \mh 2  \mh 3  \mh 4$  & 25.97M & 53.68\% & 41.05\% & 29.21\% & 11.92\% \\
\midrule
  \multicolumn{6}{c}{\textit{5-stage Network}}\\
\midrule
    $D \mh 1 \mh 1 \mh 3 \mh 1 \mh 2$ & 25.42M & 48.85\% & 36.89\% & 25.98\% & 10.37\% \\
    $D \mh 1 \mh 1 \mh 3 \mh 2 \mh 1$ & 25.42M & 50.14\% & 37.33\% & 26.11\% & 10.35\% \\
    $D \mh 3 \mh 6 \mh 2 \mh 2 \mh 1$ & 25.85M & 51.64\% & 39.12\% & 28.23\% & 12.24\% \\
    $D \mh 2 \mh 3 \mh 7 \mh 1 \mh 1$ & 26.06M & 52.16\% & 39.79\% & 28.40\% & 11.72\% \\
    $D \mh 3 \mh 4 \mh 6 \mh 2 \mh 1$ & 29.76M & 53.67\% & 41.25\% & 29.88\% & 12.65\% \\
\midrule
  \multicolumn{6}{c}{\textit{6-stage Network}}\\
\midrule
    $D\mh 1\mh 1\mh 1\mh 1\mh 1\mh 1$ & 27.39M & 40.82\% & 29.46\% & 20.00\% & 7.52\% \\
\bottomrule
\end{tabular}
\label{tab:full-depth-stage}
\end{table}

\section{Roadmap Results}
\label{sec:appendix-roadmap}

This section presents detailed results for the roadmap we take to construct the \ac{ra} family using Table \ref{tab:full-roadmap}. 
We demonstrate each architecture component in the cumulative \ac{ra} construction process improves natural and \ac{pgd}10-4 in the main paper. 
In Table \ref{tab:full-roadmap}, we show that the accuracy gain is also consistent on \ac{pgd}10-2 and \ac{pgd}10-8. 

\begin{table}
\small
\centering
\caption{\ac{pgd}10 robustness of all stage-level designs. Bold font means the results have been presented in the paper. All configurations are trained with Fast-\ac{at} and evaluated on full ImageNet validation set. ResNet-50 serves as the baseline. We presented ``Stem width 96'' and ``Move down ($\downarrow$) downsampling'' for the stem stage, and ``Dense ratio 2'' for the dense connection in the main paper. We complete the results by providing all other configurations, and \ac{pgd} attack budgets here.}
\begin{tabular}{@{}l@{\hspace{1.8mm}}r@{\hspace{1.3mm}}r@{\hspace{1.5mm}}r@{\hspace{1.5mm}}r@{\hspace{1.5mm}}r@{}}
\toprule
    Config & \#Param & Natural & PGD10$\mh$2 & PGD10$\mh$4 & PGD10$\mh$8 \\
\midrule
    ResNet-50 & 25.56M & 56.09\% & 42.66\% & 30.43\% & 12.61\% \\
\midrule
  \multicolumn{6}{c}{\textit{Stem Stage}}\\
\midrule
Stem width 32             & 25.54M & 55.89\% & 41.64\% & 29.73\% & 13.25\% \\
Stem width 96             & 25.57M & \textbf{57.29\%} & 44.55\% & \textbf{32.06\%} & 13.74\% \\
Stem kernel 3             & 25.55M & 38.93\% &  0.46\% &  0.55\% &  0.30\% \\
Stem kernel 5             & 25.55M & 59.59\% &  0.38\% &  0.09\% &  0.04\% \\
Stem kernel 9             & 25.56M & 55.75\% & 43.00\% & 31.19\% & 13.63\% \\
Move down ($\downarrow$)  & \multirow{2}{*}{25.56M} & \multirow{2}{*}{\textbf{57.08\%}} & \multirow{2}{*}{45.19\%} & \multirow{2}{*}{\textbf{33.08\%}} & \multirow{2}{*}{14.50\%} \\
downsampling              & & & & & \\
Downsample                & \multirow{2}{*}{25.56M} & \multirow{2}{*}{56.03\%} & \multirow{2}{*}{44.48\%} & \multirow{2}{*}{32.86\%} & \multirow{2}{*}{14.71\%} \\
factor 2                  & & & & & \\
``Patchify 4''                         & 25.55M & 55.40\% & 43.45\% & 31.68\% & 13.80\% \\
``Patchify 2''                         & 25.55M & 56.38\% & 44.21\% & 31.91\% & 13.48\% \\
\midrule
  \multicolumn{6}{c}{\textit{Dense Connection}}\\
\midrule
Dense ratio 2    & 25.56M & \textbf{55.93\%} & 42.85\% & \textbf{30.73\%} & 12.67\% \\
Dense ratio 3    & 25.56M & 53.45\% & 40.70\% & 29.39\% & 12.84\% \\
Dense ratio 4    & 25.56M & 55.02\% & 42.44\% & 30.52\% & 12.98\% \\
Dense ratio 5    & 25.56M & 54.45\% & 41.96\% & 30.07\% & 12.49\% \\
Dense ratio 5    & \multirow{2}{*}{25.56M} & \multirow{2}{*}{49.68\%} & \multirow{2}{*}{37.32\%} & \multirow{2}{*}{26.15\%} & \multirow{2}{*}{10.28\%} \\
\ac{relu}$\mh$\ac{relu}$\mh$0 & & & & & \\
\bottomrule
\end{tabular}
\label{tab:full-stage}
\end{table}

\begin{table}
\small
\centering
\caption{\ac{pgd}10 robustness of all block-level designs. Bold font means the results have been presented in the paper. All configurations are trained with Fast-\ac{at} and evaluated on full ImageNet validation set. ResNet-50 serves as the baseline. Bold means results have already appeared in the main paper. We complete the results by providing all other configurations and \ac{pgd} attack budgets here. For activation, 0$\mh$0$\mh$\ac{relu} means only the last activation layer is preserved in a block and the first two are discarded. The same also applies to normalization.}
\begin{tabular}{@{}l@{\hspace{1.5mm}}r@{\hspace{1.3mm}}r@{\hspace{1.5mm}}r@{\hspace{1.5mm}}r@{\hspace{1.5mm}}r@{}}
\toprule
    Config & \#Param & Natural & PGD10$\mh$2 & PGD10$\mh$4 & PGD10$\mh$8 \\
\midrule
    ResNet-50 & 25.56M & 56.09\% & 42.66\% & 30.43\% & 12.61\% \\
\midrule
  \multicolumn{6}{c}{\textit{Kernel Size}}\\
\midrule
Kernel size 5 & 45.68M & \textbf{56.73\%} & 44.55\% & \textbf{32.77\%} & 14.62\% \\
Kernel size 7 & 75.86M & \textbf{59.70\%} & 47.28\% & \textbf{34.67\%} & 14.99\% \\
\midrule
  \multicolumn{6}{c}{\textit{Dilation}}\\
\midrule
Dilation 2 & 25.56M & \textbf{52.98\%} & 40.38\% & \textbf{28.38\%} & 11.79\% \\
Dilation 3 & 25.56M & \textbf{52.10\%} & 39.69\% & \textbf{27.97\%} & 11.05\% \\
\midrule
  \multicolumn{6}{c}{\textit{Activation}}\\
\midrule
Act. \ac{gelu}                    & 25.56M & \textbf{57.48\%} & 45.05\% & \textbf{33.12\%} & 14.80\% \\
Act. \ac{silu}                    & 25.56M & \textbf{58.19\%} & 46.21\% & \textbf{34.07\%} & 14.68\% \\
Act. \ac{prelu}                   & 25.56M & 55.81\% & 42.52\% & 30.38\% & 12.76\% \\
Act. \ac{psilu}                   & 25.56M & \textbf{56.38\%} & 44.90\% & \textbf{33.76\%} & 15.40\% \\
Act. \ac{pssilu}                  & 25.56M & 57.43\% & 44.44\% & 32.22\% & 13.71\% \\
\ac{relu}$\mh$\ac{relu}$\mh$0     & 25.56M & 51.54\% & 38.69\% & 27.05\% & 10.94\% \\
\ac{relu}$\mh$0$\mh$\ac{relu}     & 25.56M & 53.91\% & 41.22\% & 29.62\% & 12.30\% \\
0$\mh$\ac{relu}$\mh$\ac{relu}     & 25.56M & 54.81\% & 42.10\% & 30.34\% & 12.86\% \\
0$\mh$0$\mh$\ac{relu}             & 25.56M & 51.03\% & 39.12\% & 28.15\% & 12.09\% \\
0$\mh$\ac{relu}$\mh$0             & 25.56M & 47.18\% & 34.85\% & 24.12\% &  9.51\% \\
\ac{relu}$\mh$0$\mh$0             & 25.56M & 44.21\% & 32.34\% & 22.24\% &  8.77\% \\
\midrule
  \multicolumn{6}{c}{\textit{\acf{se}}}\\
\midrule
SE (\ac{relu})    & 27.73M & \textbf{57.83\%} & 45.09\% & \textbf{32.64\%} & 14.01\% \\
SE (\ac{silu})    & 27.73M & 58.49\% & 45.79\% & 33.63\% & 14.51\% \\
SE (\ac{gelu})    & 27.73M & 58.27\% & 45.66\% & 33.55\% & 14.56\% \\
SE (\ac{psilu})   & 27.73M & 56.98\% & 44.19\% & 32.19\% & 13.68\% \\
SE (\ac{pssilu})  & 27.73M & 57.55\% & 45.27\% & 33.33\% & 14.73\% \\
\midrule
  \multicolumn{6}{c}{\textit{Normalization}}\\
\midrule
Norm. \ac{in}              & 25.51M & 17.15\% & 12.49\% &  8.54\% &  3.55\% \\
\ac{bn}$\mh$\ac{bn}$\mh$0  & 25.53M & \textbf{54.15\%} & 41.12\% & \textbf{29.59\%} & 12.36\% \\
\ac{bn}$\mh$0$\mh$\ac{bn}  & 25.55M & \textbf{56.04\%} & 43.29\% & \textbf{31.34\%} & 13.37\% \\
0$\mh$\ac{bn}$\mh$\ac{bn}  & 25.55M & \textbf{56.18\%} & 43.64\% & \textbf{31.61\%} & 13.47\% \\
0$\mh$0$\mh$\ac{bn}        & 25.54M & 54.47\% & 41.91\% & 30.13\% & 12.65\% \\
0$\mh$\ac{bn}$\mh$0        & 25.52M & 54.55\% & 41.94\% & 30.06\% & 12.62\% \\
\ac{bn}$\mh$0$\mh$0        & 25.52M & 54.44\% & 41.47\% & 29.72\% & 12.50\% \\
\bottomrule
\end{tabular}
\label{tab:full-block}
\end{table}

\begin{table*}
\small
  \centering
  \caption{The roadmap outlines the path we take to cumulatively improve the robustness and construct \ac{ra}-S (${\sim}26$M), \ac{ra}-M (${\sim}46$M), and \ac{ra}-L (${\sim}104$M) based on our guidelines. 
  Natural and \ac{pgd}10-4 accuracies were already shown in the main paper. 
  \ac{pgd}10-2 and \ac{pgd}10-8 show similar trends of accuracy improvement as \ac{pgd}10-4. }
  \begin{tabular}{l@{\hspace{5mm}}|l@{\hspace{5mm}}r@{\hspace{8mm}}r@{\hspace{5mm}}r@{\hspace{5mm}}r@{\hspace{5mm}}r}
  \toprule
  & Configurations & \#Param & Natural & \ac{pgd}10-2 & \ac{pgd}10-4 & \ac{pgd}10-8 \\
  \midrule
  \multicolumn{7}{c}{\textit{Small}: ResNet-50 $\rightarrow$ \ac{ra}-S ($\mathcal{S}_{7}$)} \\
  \midrule
$\mathcal{S}_0$ & ResNet-50                                                                    & 25.71M & 56.09\% & 42.66\% & 30.43\% & 12.61\% \\
$\mathcal{S}_1$ & $\mathcal{S}_0$ + $D\mh 5\mh 8\mh 13\mh 1$                                   & 25.56M & 57.35\% & 44.83\% & 33.33\% & 15.46\% \\
$\mathcal{S}_{2a}$ & $\mathcal{S}_1$ + $g=2, e=2, b=0.25$                                      & 25.84M & 57.98\% & 46.00\% & 33.94\% & 15.27\% \\
$\mathcal{S}_{2b}$ & $\mathcal{S}_1$ + $g=1, e=1.5, b=0.25$                                    & 25.53M & 57.52\% & 44.60\% & 32.83\% & 14.23\% \\
$\mathcal{S}_{3}$ & $\mathcal{S}_{2a}$ + Stem width 96 + Move down ($\downarrow$) downsampling & 25.85M & 57.82\% & 46.37\% & 34.86\% & 15.92\% \\
$\mathcal{S}_{4}$ & $\mathcal{S}_{3}$ + \ac{se} (\ac{relu})                                    & 26.15M & 60.57\% & 49.05\% & 36.61\% & 16.43\% \\
$\mathcal{S}_{5}$ & $\mathcal{S}_{4}$ + Act. \ac{silu}                                         & 26.15M & 62.04\% & 51.41\% & 39.48\% & 18.95\% \\
$\mathcal{S}_{6}$ & $\mathcal{S}_{5}$ + \ac{se} (\ac{silu})                                    & 26.15M & 60.32\% & 49.74\% & 38.24\% & 18.18\% \\
$\mathcal{S}_{7}$ & $\mathcal{S}_{5}$ + Norm$\mh$0$\mh$BN$\mh$BN                               & 26.14M & 62.27\% & 51.67\% & 39.88\% & 18.99\% \\
  \midrule
  \multicolumn{7}{c}{\textit{Medium}: \ac{ra}-S ($\mathcal{S}_{7}$) $\rightarrow$ \ac{ra}-M ($\mathcal{M}_{2}$)} \\
  \midrule
$\mathcal{M}_{1}$ & $\mathcal{S}_{7}$ + Kernel size 5                      & 45.95M & 63.82\% & 52.89\% & 41.00\% & 19.90\% \\
$\mathcal{M}_{2}$ & $\mathcal{S}_{7}$ + $D\mh 7\mh 11\mh 18\mh 1$          & 45.90M & 64.40\% & 53.97\% & 42.06\% & 20.98\% \\
$\mathcal{M}_{3}$ & $\mathcal{S}_{7}$ + $W\mh 384\mh 760\mh 1504\mh 2944$  & 46.16M & 63.52\% & 53.11\% & 41.43\% & 20.27\% \\
  \midrule
  \multicolumn{7}{c}{\textit{Large}: \ac{ra}-M ($\mathcal{M}_{2}$) $\rightarrow$ \ac{ra}-L ($\mathcal{L}_{2}$)} \\
  \midrule
$\mathcal{L}_{1}$ & $\mathcal{M}_{2}$ + Kernel size 7                         & 103.89M & 64.08\% & 52.92\% & 40.70\% & 19.61\% \\
$\mathcal{L}_{2}$ & $\mathcal{M}_{2}$ + $W\mh 512\mh 1024\mh 2016\mh 4032$    & 104.07M & 66.08\% & 55.52\% & 43.81\% & 22.50\% \\
$\mathcal{L}_{3}$ & $\mathcal{M}_{2}$ + $D\mh 8\mh 13\mh 21\mh 2$             & 104.13M & 64.91\% & 54.64\% & 43.09\% & 21.81\% \\
$\mathcal{L}_{4}$ & $\mathcal{M}_{2}$ + $D\mh 10\mh 16\mh 26\mh 2$            & 104.14M & 65.28\% & 54.49\% & 42.85\% & 21.42\% \\
  \bottomrule
  \end{tabular}
  \label{tab:full-roadmap}
\end{table*}

\section{SOTA Architecture Comparisons}
\label{sec:appendix-sota}

\subsection{Fast-AT Comparisons}
\label{sec:appendix-fastat}

This section presents the detailed results of \ac{ra} and other \ac{sota} architectures after Fast-\ac{at} using Table \ref{tab:full-fat}. 
With a similar model capacity, \ac{ra}-S outperforms ResNet-50 and ResNeXt-50 43$\times$4d, and \ac{ra}-M outperforms ResNet-101. 
Compared to models with larger parameters, \ac{ra}-S is even more robust than WideResNet101-2 despite having $4.85\times$ fewer parameters. 
The accuracy continues to increase while scaling up the \ac{ra} models, with \ac{ra}-L achieving the highest natural and adversarial accuracies. 

\subsection{Standard-AT Comparisons}
\label{sec:appendix-aa}

This section compares our \acp{ra} with other \ac{sota} models against both \ac{pgd} and \ac{aa} in Table \ref{tab:full-sota-aa}. 
For \ac{aa}, all three \acp{ra} outperform their \ac{xcit} counterparts.
Using the same training configurations as Salman \etal \cite{salman2020adversarially}, \ac{ra}-S surpasses ResNet-50 \ac{aa} accuracy by $9.18$ percentage points, and is even more robust than WideResNet50-2 with $2.6\times$ fewer parameters. 
The robustness continues to scale with model capacity, and \ac{ra}-L achieves the new \ac{sota} \ac{aa} accuracy on the RobustBench leaderboard. 
It is important to note that ResNet-50+DiffPure \cite{nie2022diffusion} designed a novel \ac{at} method via using diffusion models 
\cite{ho2020denoising} 
for adversarial purification. 
Although the method improves the \ac{aa} accuracy by $5.97$ percentage points, our architecture modifications show stronger robustness even without finetuning the Standard-\ac{at} method. 
We believe a carefully designed training recipe can further improve \acp{ra}' robustness. 
For \ac{pgd}, the \ac{ra}-S again outperforms ResNet-50 and even WideResNet50-2 using the same Standard-\ac{at} configurations. 
Overall, our \acp{ra} outperform both \textit{ConvNets} and \textit{Transformers} with similar total parameters. 

\begin{table*}[t]
\small
\centering
\caption{Our \ac{ra} model outperforms \textit{ConvNets} and \textit{Transformers} with similar total parameters against $\ell_\infty = 4 / 255$ \ac{aa} and $\ell_\infty = 2 / 255, 4 / 255, 8 / 255$ \ac{pgd} attacks. 
Using the same training configurations as Salman \etal \cite{salman2020adversarially}, our model outperforms both ResNet-50 and WideResNet50-2. 
Every \ac{ra} model outperforms its \ac{xcit} counterpart at a similar capacity. 
}
\begin{tabular}{lrrrrrrrr}
\toprule
    Architecture & \#Param & Natural & \ac{aa} & \ac{pgd}10-4 & \ac{pgd}50-4 & \ac{pgd}100-4 & \ac{pgd}100-2 & \ac{pgd}100-8 \\
\midrule
    ResNet-18 \cite{salman2020adversarially}   & 12M & 52.49\% & 25.32\% & 30.06\% & 29.61\% & 29.61\% & 40.98\% & 11.57\% \\
    RobNet-large \cite{guo2020meets}           & 13M & 61.26\% & -     & 37.16\% & 37.15\% & 37.14\% & - & - \\
    PoolFormer-M12 \cite{debenedetti2022light} & 22M & 66.16\% & 34.72\% & -     & - & - & - & - \\ 
    DeiT-S \cite{bai2021transformers}          & 22M & 66.50\% & 35.50\% & 41.03\% & 40.34\% & 40.32\% & - & - \\
    DeiT-S+DiffPure \cite{nie2022diffusion}    & 22M & \textbf{73.63\%} & 43.18\% & - & - & - & - & -  \\
    ResNet-50 \cite{salman2020adversarially}   & 26M & 63.87\% & 34.96\% & 39.66\% & 38.98\% & 38.96\% & 52.15\% & 15.83\% \\
    ResNet-50+DiffPure \cite{nie2022diffusion} & 26M & 67.79\% & 40.93\% & - & - & - & - & - \\
    ResNet50+SiLU \cite{xie2020smooth}         & 26M & 69.70\% & - & 43.00\% & 41.90\% & - & - & - \\
    ResNet50+GELU \cite{bai2021transformers}   & 26M & 67.38\% & 35.51\% & 40.98\% & 40.28\% & 40.27\% & - & - \\
    ResNet-50-R \cite{huang2021exploring}      & 26M & 56.63\% & - & - & 31.14\% & - & - & - \\
    XCiT-S12 \cite{debenedetti2022light}       & 26M & 72.34\% & 41.78\% & - & - & - & - & - \\
    \ac{ra}-S & 26M & 70.17\% & \textbf{44.14\%}   & \textbf{48.19\%} & \textbf{47.78\%} & \textbf{47.77\%} & \textbf{60.06\%} & \textbf{21.77\%} \\
\midrule
    XCiT-M12 \cite{debenedetti2022light}       & 46M & \textbf{74.04\%} & 45.24\% & - & - & - & - & - \\
    \ac{ra}-M                                  & 46M & 71.88\% & \textbf{46.26\%} & \textbf{49.84\%} & \textbf{49.32\%} & \textbf{49.30\%} & \textbf{61.89\%} & \textbf{23.01\%} \\
\midrule
    WideResNet50-2 \cite{salman2020adversarially} & 69M & 68.41\% & 38.14\% & 42.51\% & 41.33\% & 41.24\% & 55.86\% & 16.29\% \\
    WideResNet50-2+DiffPure \cite{nie2022diffusion} & 69M & 71.16\% & 44.39\% & - & - & - & - & - \\
    Swin-B \cite{mo2022adversarial}            & 88M & 74.36\% & 38.61\% & - & - & - & - & -\\
    XCiT-L12 \cite{debenedetti2022light}       & 104M & \textbf{73.76\%} & 47.60\% & - & - & - & - & - \\
    \ac{ra}-L                                  & 104M & 73.44\% & \textbf{48.94\%} & \textbf{51.72\%} & \textbf{51.04\%} & \textbf{51.03\%} & \textbf{63.49\%} & \textbf{25.31\%} \\
\bottomrule
\end{tabular}
\label{tab:full-sota-aa}
\end{table*}

\begin{table*}
\small
\centering
\caption{\ac{pgd}10 robustness of width. Bold font means the results have been presented in the paper. All configurations are trained with Fast-\ac{at} and evaluated on full ImageNet validation set. ResNet-50 serves as the baseline. In the main paper, we presented $BM\mh0.5\mh0.5\mh0.5\mh0.5$ and $BM\mh0.5\mh0.5\mh0.25\mh0.25$ for bottleneck multiplier, $G\mh2\mh2\mh2\mh2$ for group convolution groups, $W\mh512\mh768\mh1152\mh1728$ for expansion ratio, and the combined model. We complete the results by providing all other configurations, and \ac{pgd} attack budgets here. }
\begin{tabular}{@{}l@{\hspace{3mm}}l@{\hspace{3mm}}l@{\hspace{3mm}}r@{\hspace{2mm}}r@{\hspace{2mm}}r@{\hspace{2mm}}r@{\hspace{2mm}}r@{}}
\toprule
    Channel & Group & Bottleneck Multiplier & \#Param & Natural & PGD10$\mh$2 & PGD10$\mh$4 & PGD10$\mh$8 \\
\midrule
    \multicolumn{3}{c}{ResNet-50} & 25.56M & 56.09\% & 42.66\% & 30.43\% & 12.61\% \\
\midrule
  \multicolumn{8}{c}{\textit{Bottleneck Multiplier}}\\
\midrule
$W\mh320\mh672\mh1456\mh3136$ & $G\mh1\mh1\mh1\mh1$ & $BM\mh0.125\mh0.125\mh0.125\mh0.125$ & 25.47M & 53.47\% & 41.42\% & 30.11\% & 13.40\% \\
$W\mh128\mh256\mh568\mh1304$  & $G\mh1\mh1\mh1\mh1$ & $BM\mh0.5\mh0.5\mh0.5\mh0.5$         & 25.57M & \textbf{55.31\%} & 42.48\% & \textbf{30.52\%} & 13.23\% \\
$W\mh64\mh144\mh320\mh720$    & $G\mh1\mh1\mh1\mh1$ & $BM\mh1\mh1\mh1\mh1$                 & 25.61M & 53.07\% & 40.93\% & 29.54\% & 12.70\% \\
$W\mh32\mh72\mh168\mh384$     & $G\mh1\mh1\mh1\mh1$ & $BM\mh2\mh2\mh2\mh2$                 & 25.72M & 51.17\% & 38.79\% & 27.32\% & 11.22\% \\
$W\mh16\mh32\mh88\mh200$      & $G\mh1\mh1\mh1\mh1$ & $BM\mh4\mh4\mh4\mh4$                 & 26.19M & 47.67\% & 35.93\% & 25.30\% & 10.32\% \\
$W\mh256\mh512\mh168\mh384$   & $G\mh1\mh1\mh1\mh1$ & $BM\mh0.25\mh0.25\mh2\mh2$           & 26.42M & 52.33\% & 39.79\% & 28.52\% & 12.30\% \\
$W\mh24\mh48\mh1024\mh2048$   & $G\mh1\mh1\mh1\mh1$ & $BM\mh4\mh4\mh0.25\mh0.25$           & 25.20M & 55.78\% & 43.09\% & 30.79\% & 12.89\% \\
$W\mh128\mh256\mh1024\mh2048$ & $G\mh1\mh1\mh1\mh1$ & $BM\mh0.5\mh0.5\mh0.25\mh0.25$       & 24.83M & \textbf{56.11\%} & 43.38\% & \textbf{31.26\%} & 13.47\% \\
\midrule
  \multicolumn{8}{c}{\textit{Group Convolution Groups}}\\
\midrule
$W\mh256\mh512\mh1080\mh2504$ & $G\mh2\mh2\mh2\mh2$        & $BM\mh0.25\mh0.25\mh0.25\mh0.25$ & 26.02M & \textbf{57.31\%} & 44.25\% & \textbf{32.09\%} & 13.91\% \\
$W\mh288\mh576\mh1248\mh2592$ & $G\mh4\mh4\mh4\mh4$        & $BM\mh0.25\mh0.25\mh0.25\mh0.25$ & 25.58M & 56.28\% & 44.00\% & 31.52\% & 13.33\% \\
$W\mh256\mh512\mh1280\mh2816$ & $G\mh8\mh8\mh8\mh8$        & $BM\mh0.25\mh0.25\mh0.25\mh0.25$ & 25.81M & 56.54\% & 42.49\% & 30.07\% & 12.86\% \\
$W\mh256\mh576\mh1344\mh2816$ & $G\mh16\mh16\mh16\mh16$    & $BM\mh0.25\mh0.25\mh0.25\mh0.25$ & 25.61M & 54.83\% & 42.92\% & 31.03\% & 13.28\% \\
$W\mh304\mh640\mh1384\mh2848$ & $G\mh76\mh160\mh337\mh712$ & $BM\mh0.25\mh0.25\mh0.25\mh0.25$ & 25.52M & 55.17\% & 42.34\% & 30.45\% & 12.72\% \\
$W\mh256\mh512\mh1040\mh2112$ & $G\mh8\mh8\mh1\mh1$        & $BM\mh0.25\mh0.25\mh0.25\mh0.25$ & 26.13M & 55.49\% & 42.42\% & 30.78\% & 12.82\% \\
$W\mh256\mh512\mh1248\mh2784$ & $G\mh1\mh1\mh8\mh8$        & $BM\mh0.25\mh0.25\mh0.25\mh0.25$ & 25.69M & 55.94\% & 43.28\% & 31.15\% & 13.92\% \\
$W\mh256\mh512\mh1248\mh2592$ & $G\mh2\mh2\mh4\mh4$        & $BM\mh0.25\mh0.25\mh0.25\mh0.25$ & 25.41M & 57.13\% & 43.88\% & 31.44\% & 13.48\% \\
\midrule
  \multicolumn{8}{c}{\textit{Channel / Expansion Ratio}}\\
\midrule
$W\mh1112\mh1112\mh1112\mh1112$ & $G\mh1\mh1\mh1\mh1$ & $BM\mh0.25\mh0.25\mh0.25\mh0.25$ & 25.70M & 56.77\% & 43.18\% & 31.08\% & 13.68\% \\
$W\mh512\mh768\mh1152\mh1728$   & $G\mh1\mh1\mh1\mh1$ & $BM\mh0.25\mh0.25\mh0.25\mh0.25$ & 25.95M & \textbf{57.17\%} & 44.05\% & \textbf{32.04\%} & 14.06\% \\
$W\mh144\mh360\mh904\mh2264$    & $G\mh1\mh1\mh1\mh1$ & $BM\mh0.25\mh0.25\mh0.25\mh0.25$ & 26.01M & 53.89\% & 41.83\% & 30.33\% & 13.38\% \\
$W\mh88\mh264\mh792\mh2376$     & $G\mh1\mh1\mh1\mh1$ & $BM\mh0.25\mh0.25\mh0.25\mh0.25$ & 25.81M & 52.39\% & 40.60\% & 29.36\% & 12.58\% \\
\midrule
  \multicolumn{8}{c}{\textit{Combined}}\\
\midrule
$W\mh512\mh768\mh1152\mh1728$   & $G\mh2\mh2\mh2\mh2$ & $BM\mh0.5\mh0.5\mh0.25\mh0.25$ & 24.43M & \textbf{56.64\%} & 43.56\% & \textbf{31.04\%} & 13.17\% \\

\bottomrule
\end{tabular}
\label{tab:full-width}
\end{table*}

\begin{table*}
\small
\centering
\caption{\ac{pgd}10 robustness of combining depth and width. We use a bold font to highlight results that have been presented in the paper. Specifically,
the paper uses a scatter plot to visualize how the \ac{pgd}10-4 accuracy changes  as we vary depth and width.
Here, we additionally show the results for \ac{pgd}10-2 and \ac{pgd}10-8.
All configurations are trained with Fast-\ac{at} and evaluated on full ImageNet validation set. 
}
\begin{tabular}{@{}llr@{\hspace{2mm}}r@{\hspace{2mm}}r@{\hspace{2mm}}r@{\hspace{2mm}}r@{}}
\toprule
    Depth & Width & \#Param & Natural & PGD10$\mh$2 & PGD10$\mh$4 & PGD10$\mh$8 \\
\midrule
$D\mh1\mh2\mh4\mh1$     & $W\mh768\mh1152\mh1712\mh2560$ & 25.69M & 54.28\% & 41.16\% & \textbf{29.10\%} & 11.83\% \\
$D\mh2\mh4\mh7\mh1$     & $W\mh648\mh968\mh1456\mh2160$  & 25.55M & 57.25\% & 43.60\% & \textbf{31.52\%} & 13.59\% \\
$D\mh4\mh6\mh10\mh1$    & $W\mh576\mh848\mh1280\mh1904$  & 25.51M & 57.08\% & 44.18\% & \textbf{32.32\%} & 14.46\% \\
$D\mh5\mh8\mh13\mh1$    & $W\mh512\mh768\mh1152\mh1728$  & 25.18M & 57.24\% & 44.69\% & \textbf{33.05\%} & 15.36\% \\
$D\mh8\mh12\mh20\mh2$   & $W\mh424\mh632\mh944\mh1416$   & 25.37M & 57.74\% & 44.79\% & \textbf{33.15\%} & 14.87\% \\
$D\mh10\mh16\mh26\mh2$  & $W\mh376\mh568\mh856\mh1280$   & 25.56M & 61.36\% & 44.92\% & \textbf{27.23\%} &  5.67\% \\
$D\mh20\mh32\mh52\mh4$  & $W\mh272\mh416\mh616\mh928$    & 25.52M & 55.76\% & 43.28\% & \textbf{31.31\%} & 13.03\% \\

\bottomrule
\end{tabular}
\label{tab:full-depth-width}
\end{table*}